\documentclass[acmtog]{acmart}
\AtBeginDocument{%
  \providecommand\BibTeX{{%
    \normalfont B\kern-0.5em{\scshape i\kern-0.25em b}\kern-0.8em\TeX}}}

\usepackage{colortbl}
\usepackage{siunitx}
\usepackage{xfrac}
\usepackage{accents}
\usepackage{paralist}

\usepackage{booktabs} %
\usepackage{xspace}
\usepackage{array}
\usepackage{hyperref}
\usepackage{dblfloatfix}

\usepackage{lipsum}
\usepackage{subcaption}

\usepackage{wrapfig}

\usepackage{amsmath}
\usepackage{bm}

\usepackage{siunitx}
\usepackage{enumitem}
\setlist{nosep} %

\newcolumntype{L}[1]{>{\raggedright\let\newline\\\arraybackslash\hspace{0pt}}m{#1}}
\newcolumntype{C}[1]{>{\centering\let\newline\\\arraybackslash\hspace{0pt}}m{#1}}
\newcolumntype{R}[1]{>{\raggedleft\let\newline\\\arraybackslash\hspace{0pt}}m{#1}}

\newcommand{\sect}[1]{Section~\ref{#1}}

\newcommand{\fig}[1]{Figure~\ref{#1}}
\newcommand{\tbl}[1]{Table~\ref{#1}}

\newcommand{\degree}{\ensuremath{^\circ}\xspace}
\newcommand{\ignore}[1]{}

\def\around{{\raise.17ex\hbox{$\scriptstyle\mathtt{\sim}$}}}

\makeatletter
\DeclareRobustCommand\onedot{\futurelet\@let@token\@onedot}
\def\@onedot{\ifx\@let@token.\else.\null\fi\xspace}
\def\eg{e.g\onedot} 
\def\ie{i.e\onedot} 
 
\def\etc{etc\onedot} \def\vs{vs\onedot}
\def\wrt{w.r.t\onedot}

\makeatother

\def\to{$\,\to\,$}

\definecolor{MyLightGreen}{rgb}{0.92,1,0.95}
\definecolor{MyLightBlue}{rgb}{0.92,0.95,1}
\definecolor{MyDarkBlue}{rgb}{0,0.08,1}
\definecolor{MyDarkGreen}{rgb}{0.02,0.6,0.02}
\definecolor{MyDarkCyan}{rgb}{0,0.7,0.7}
\definecolor{MyDarkOrange}{rgb}{0.40,0.2,0.02}
\definecolor{MyPurple}{RGB}{111,0,255}
\definecolor{MyRed}{rgb}{1.0,0.0,0.0}
\definecolor{MyDarkRed}{rgb}{0.8,0,0}
\definecolor{MyGold}{rgb}{0.75,0.6,0.12}
\definecolor{MyDarkgray}{rgb}{0.66, 0.66, 0.66}
\definecolor{JonYellow}{rgb}{1,1, 0.6}
\definecolor{JonOrange}{rgb}{1, 0.8, 0.6}
\definecolor{JonRed}{rgb}{1, 0.6, 0.6}

\newcommand{\modelfull}{Neural Radiance Factorization (NeRFactor)\xspace}
\newcommand{\model}{NeRFactor\xspace}
\newcommand{\modelmicrofacet}{using microfacet\xspace}
\newcommand{\modelnerfshape}{using NeRF's shape\xspace}

\newcommand{\rev}[1]{#1}

\newcommand{\dummytext}{}

\newcommand{\www}{\url{people.csail.mit.edu/xiuming/projects/nerfactor/}\xspace}
\newcommand{\suppvideo}{\href{https://www.youtube.com/watch?v=UUVSPJlwhPg}{the supplemental video}\xspace}
\newcommand{\repo}{\href{https://github.com/google/nerfactor}{our GitHub repository}\xspace} 
\setcopyright{rightsretained}
\acmJournal{TOG}
\acmYear{2021}\acmVolume{40}\acmNumber{6}\acmArticle{237}\acmMonth{12}
\acmDOI{10.1145/3478513.3480496}

\setcitestyle{square,nosort}

\begin{document}

\title[NeRFactor]{NeRFactor: Neural Factorization of Shape and Reflectance Under an Unknown Illumination}

\author{Xiuming Zhang}
\email{xiuming@csail.mit.edu}
\orcid{0000-0002-4326-727X}
\affiliation{%
  \institution{MIT CSAIL}
  \country{USA}
}
\author{Pratul P.\ Srinivasan}
\affiliation{%
  \institution{Google Research}
  \country{USA}
}
\author{Boyang Deng}
\affiliation{%
  \institution{Google Research}
  \country{USA}
}
\author{Paul Debevec}
\affiliation{%
  \institution{Google Research}
  \country{USA}
}
\author{William T.\ Freeman}
\affiliation{%
  \institution{MIT CSAIL \& Google Research}
  \country{USA}
}
\author{Jonathan T.\ Barron}
\affiliation{%
  \institution{Google Research}
  \country{USA}
}

\begin{abstract}

We address the problem of recovering the shape and spatially-varying reflectance of an object from multi-view images (and their camera poses) of an object illuminated by one unknown lighting condition.
This enables the rendering of novel views of the object under arbitrary environment lighting and editing of the object's material properties.
The key to our approach, which we call \modelfull, is to distill the volumetric geometry of a Neural Radiance Field (NeRF) \cite{mildenhall2020nerf} representation of the object into a surface representation and then jointly refine the geometry while solving for the spatially-varying reflectance and environment lighting.
Specifically, \model recovers 3D neural fields of surface normals, light visibility, albedo, and Bidirectional Reflectance Distribution Functions (BRDFs) without any supervision, using only a re-rendering loss, simple smoothness priors, and a data-driven BRDF prior learned from real-world BRDF measurements.
By explicitly modeling light visibility, \model is able to separate shadows from albedo and synthesize realistic soft or hard shadows under arbitrary lighting conditions.
\model is able to recover convincing 3D models for free-viewpoint relighting in this challenging and underconstrained capture setup for both synthetic and real scenes.
Qualitative and quantitative experiments show that \model outperforms classic and deep learning-based state of the art across various tasks.
Our videos, code, and data are available at \www.

\end{abstract} 
\begin{teaserfigure}
  \centering
  \includegraphics[width=\linewidth]{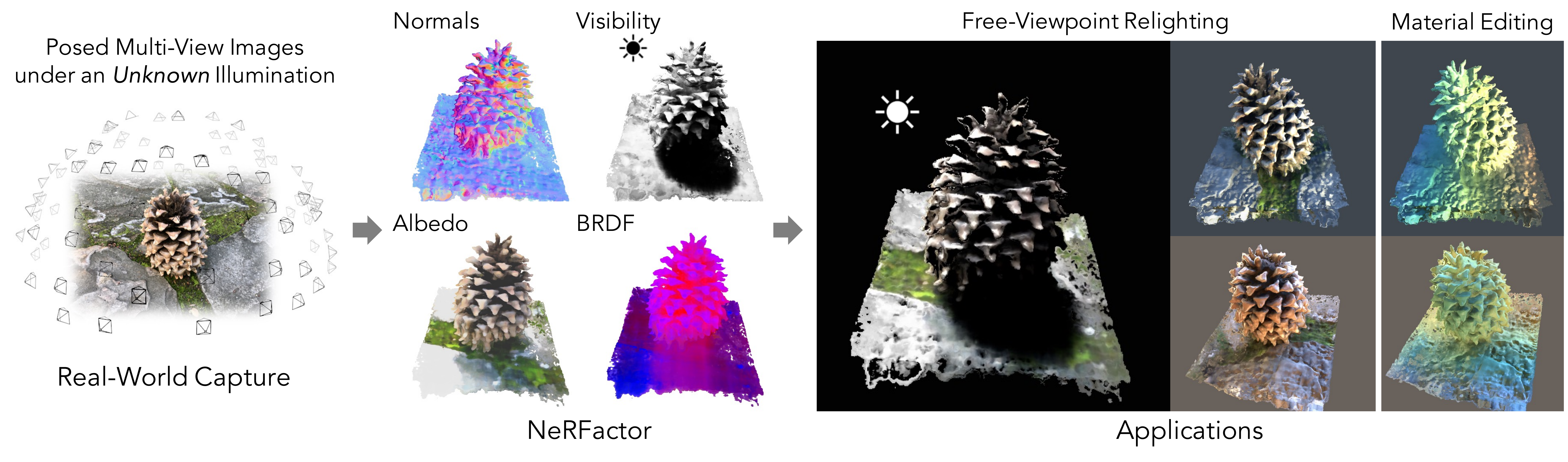}
  \vspace{-6ex}
  \caption{
  \textbf{\modelfull.}
  Given a set of multi-view images (and their camera poses) of an object lit by just one unknown illumination condition (Left), \model is able to factorize the scene appearance into 3D neural fields of surface normals, light visibility, albedo, and reflectance (Center), which enables applications such as free-viewpoint relighting that supports shadows and material editing (Right).
  }
  \label{fig:teaser}
\end{teaserfigure} 



\maketitle

\section{Introduction}

Recovering an object's geometry and material properties from captured images, such that it can be rendered from arbitrary viewpoints under novel lighting conditions, is a longstanding problem within computer vision and graphics.
The difficulty of this problem stems from its fundamentally underconstrained nature, and prior work has typically addressed this either by using additional observations such as scanned geometry, known lighting conditions, or images of the object under multiple different lighting conditions, or by making restrictive assumptions such as assuming a single material for the entire object or ignoring self-shadowing.
In this work, we demonstrate that it is possible to recover convincing relightable representations from images of an object captured under \emph{one unknown} natural illumination condition, as shown in Figure \ref{fig:teaser}.
Our key insight is that we can first optimize a Neural Radiance Field (NeRF) \cite{mildenhall2020nerf} from the input images to initialize our model's surface normals and light visibility \rev{(though we show that using Multi-View Stereo [MVS] geometry also works)}, and then jointly optimize these initial estimates along with the spatially-varying reflectance and the lighting condition to best explain the observed images.
The use of NeRF to produce a high-quality geometry estimation for initialization helps break the inherent ambiguities among shape, reflectance, and lighting, thereby allowing us to recover a full 3D model for convincing view synthesis and relighting using just a re-rendering loss, simple spatial smoothness priors for each of these components, and a novel data-driven Bidirectional Reflectance Distribution Function (BRDF) prior.
Because \model models light visibility explicitly and efficiently, it is capable of removing shadows from albedo estimation and synthesizing realistic soft or hard shadows under arbitrary novel lighting conditions.

Although the geometry estimated by NeRF is effective for view synthesis, it has two limitations that prevent it from being easily used for relighting.
First, NeRF models shape as a volumetric field, and as such it is computationally expensive to compute shading and visibility at each point along a camera ray for a full hemisphere of lighting.
Second, the geometry estimated by NeRF contains extraneous high-frequency content that, while unnoticeable in view synthesis results, introduces high-frequency artifacts into the surface normals and light visibility computed from NeRF's geometry.
We address the first issue by using a ``hard surface'' approximation of the NeRF geometry, where we only perform shading calculations at a single point along each ray, corresponding to the expected termination depth of the volume. 
We address the second issue by representing the surface normal and light visibility at any 3D location on this surface as continuous functions parameterized by Multi-Layer Perceptrons (MLPs), and encourage these functions to be close to the values derived from the pretrained NeRF and be spatially smooth.
Thus, our model, which we call \modelfull, factors the observed images into estimated environment lighting as well as a 3D surface representation of the object with surface normals, light visibility, albedo, and spatially-varying BRDFs.
This enables us to render novel views of the object under arbitrary novel environment lighting.

In summary, our main technical contributions are:
\begin{itemize}
\item a method for factorizing images of an object under an unknown lighting condition into shape, reflectance, and illumination, thereby supporting free-viewpoint relighting (with shadows) and material editing,
\item a strategy to distill NeRF-estimated volume density into surface geometry (with normals and light visibility) to use as an initialization when improving the geometry and recovering reflectance, and
\item a novel data-driven BRDF prior learned from training a latent code model on real measured BRDFs.
\end{itemize}

\paragraph{Input and output}
The input to \model is a set of multi-view images of an object illuminated under \emph{one unknown} environment lighting condition and the camera poses of these images.
\model jointly estimates a \emph{plausible} collection of surface normals, light visibility, albedo, spatially-varying BRDFs, and the environment lighting that together explain the observed views.
We then use the recovered geometry and reflectance to synthesize images of the object from novel viewpoints under arbitrary lighting.
Modeling visibility explicitly, \model is able to remove shadows from albedo and synthesize soft or hard shadows under arbitrary lighting.

\paragraph{Assumptions}
\model considers objects to be composed of hard surfaces with a single intersection point per ray, so volumetric light transport effects such as scattering, transparency, and translucency are not modeled.
In addition, we only model direct illumination to simplify computation.
Finally, our reflectance models consider materials with achromatic specular reflectance (dielectrics), so we do not model metallic materials (though one can easily extend our model for metallic materials by additionally predicting a specular color for each surface point). %
\section{Related Work}

Inverse rendering \cite{sato97,marschner98thesis,yu1999inverse,ramamoorthi01}, the task of factorizing the appearance of an object in observed images into the underlying geometry, material properties, and lighting conditions, is a longstanding problem in computer vision and graphics.
Since the full general inverse rendering problem is well-known to be severely underconstrained, most prior approaches have addressed this problem by assuming no shadow, learning priors on shape, illumination, and reflectance, or requiring additional observations such as scanned geometry, measured lighting conditions, or additional images of the object under multiple (known) lighting conditions. 

Methods for single-image inverse rendering \cite{Barron2015Shape,li2018learning,sengupta19,yu2019inverserendernet,sangsingle,wei2020object,li2020inverse,lichy2021shape} largely rely on strong priors on geometry, reflectance, and illumination learned from large datasets.
Recent methods can effectively infer plausible settings of these factors from a single image, but do not recover full 3D representations that can be viewed from arbitrary viewpoints.

Most methods that recover factorized full 3D models for relighting and view synthesis rely on additional observations instead of strong priors.
A common strategy is to use 3D geometry obtained from active scanning \cite{lensch2003image,guo2019relightables,park20chips,schmitt2020joint,nlt}, proxy models \cite{sato2003illumination,dong2014appearance,georgoulis2015gaussian,gao20,chen2020neural}, silhouette masks \cite{oxholm2014multiview,godard2015multi,xia2016recovering}, or Multi-View Stereo (MVS; followed by surface reconstruction and meshing) \cite{laffont2012rich,nam2018practical,philip2019multi,goel20} as a starting point before recovering reflectance and refined geometry.
In this work, we show that starting with geometry estimated using a state-of-the-art neural volumetric representation enables us to recover a fully-factorized 3D model just using images captured under one illumination, without requiring any additional observations.
Crucially, using initial geometry estimated in this way enables us to recover factored models for objects that have proven to be challenging for traditional geometry estimation methods, including objects with highly reflective surfaces and detailed geometry.  

A large body of work within the computer graphics community has focused on the specific subproblem of material acquisition, where the goal is to estimate Bidirectional Reflectance Distribution Function (BRDF) properties from images of materials with known (typically planar) geometry.
These methods have traditionally utilized a signal processing-based reconstruction strategy and used complex controlled camera and lighting setups to adequately sample the BRDF \cite{foo97,matusik2003,nielsen15}, and more recent methods have enabled material acquisition from more casual smartphone setups \cite{aittala15,hui17}.
However, this line of work generally requires the geometry be simple and fully known, while we focus on a more general problem where our only observations are images of an object with complex shape and spatially-varying reflectance. 

Our work builds upon a recent trend within the computer vision and graphics communities that replaces traditional shape representations such as polygon meshes or discretized voxel grids with Multi-Layer Perceptrons (MLPs) that represent geometry as parametric functions.
These MLPs are optimized to approximate continuous 3D geometry by mapping from 3D coordinates to properties of an object or scene (such as volume density, occupancy, or signed distance) at that location.
This strategy has been successful for recovering continuous 3D shape representations from 3D observations \cite{occupancynet,park2019deepsdf,tancik2020fourfeat} and from images observed under fixed lighting \cite{mildenhall2020nerf,yariv20}.
The Neural Radiance Fields (NeRF) \cite{mildenhall2020nerf} technique has been particularly successful for optimizing volumetric geometry and appearance from observed images for the purpose of rendering photorealistic novel views.

NeRF has inspired subsequent approaches that extend its neural representation to enable relighting \cite{bi2020neural,boss2020nerd,nerv,zhang2021physg}.
\rev{
We list the differences between these concurrent approaches and \model as follows.
\begin{itemize}
\item \citet{bi2020neural} and NeRV \cite{nerv} require multiple known lighting conditions, while \model handles just one unknown illumination.
\item NeRD \cite{boss2020nerd} does not model visibility or shadows, while \model does, successfully separating shadows from albedo (as will be shown).
NeRD uses an analytic BRDF, whereas \model uses a learned BRDF that encodes priors.
\item PhySG \cite{zhang2021physg} does not model visibility or shadows and uses an analytic BRDF, just like NeRD.
In addition, PhySG assumes non-spatially-varying reflectance, while \model models spatially-varying BRDFs.
\end{itemize}
}
\section{Method}
\label{sec:method}

The input to \model is assumed to be only multi-view images (and their camera poses) of an object lit by \emph{one unknown} illumination condition.
\model represents the shape and spatially-varying reflectance of an object as a set of 3D fields, each parameterized by Multi-Layer Perceptrons (MLPs) whose weights are optimized so as to ``explain'' the set of observed input images.
After optimization, \model outputs, at each 3D location $\bm{x}$ on the object's surface, the surface normal $\bm{n}$, light visibility in any direction $v(\bm{\omega_\text{i}})$, albedo $\bm{a}$, and reflectance $\bm{z_\text{BRDF}}$ that together explain the observed appearance%
\footnote{In this paper, vectors and matrices (as well as functions that return them) are in bold; scalars and scalar functions are not.}.
By recovering the object's geometry and reflectance, \model enables applications such as free-viewpoint relighting (with shadows) and material editing.

\begin{figure*}[!htbp]
\centering

\begin{subfigure}{.5\textwidth}
  \centering
  \includegraphics[width=\linewidth]{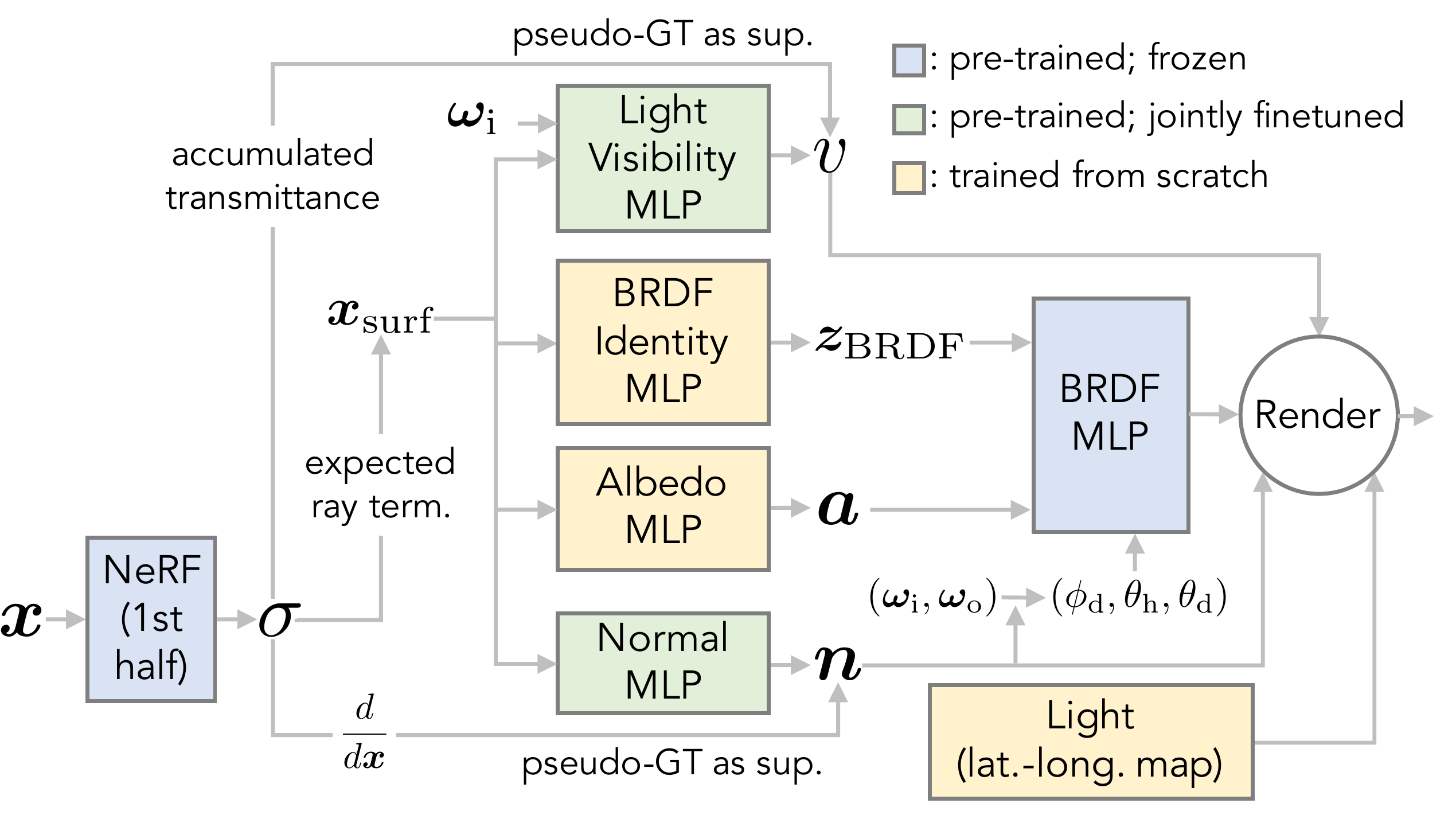}
  \caption{
  \textbf{Model.}
  \model leverages NeRF's $\sigma$-volume as an initialization to predict, for each surface location $\bm{x_\text{surf}}$, surface normal $\bm{n}$, light visibility $v$, albedo $\bm{a}$, BRDF latent code $\bm{z_\text{BRDF}}$, and the lighting condition.
  $\bm{x}$ denotes 3D locations, $\bm{\omega_\text{i}}$ light direction, $\bm{\omega_\text{o}}$ viewing direction, and $\phi_\text{d}$, $\theta_\text{h}$, $\theta_\text{d}$ Rusinkiewicz coordinates.
  Note that \model is an all-MLP architecture that models only surface points (unlike NeRF that models the entire volume).
  }
  \label{fig:model_a}
\end{subfigure}%
\hspace{1em}
\begin{subfigure}{.47\textwidth}
  \centering
  \includegraphics[width=\linewidth]{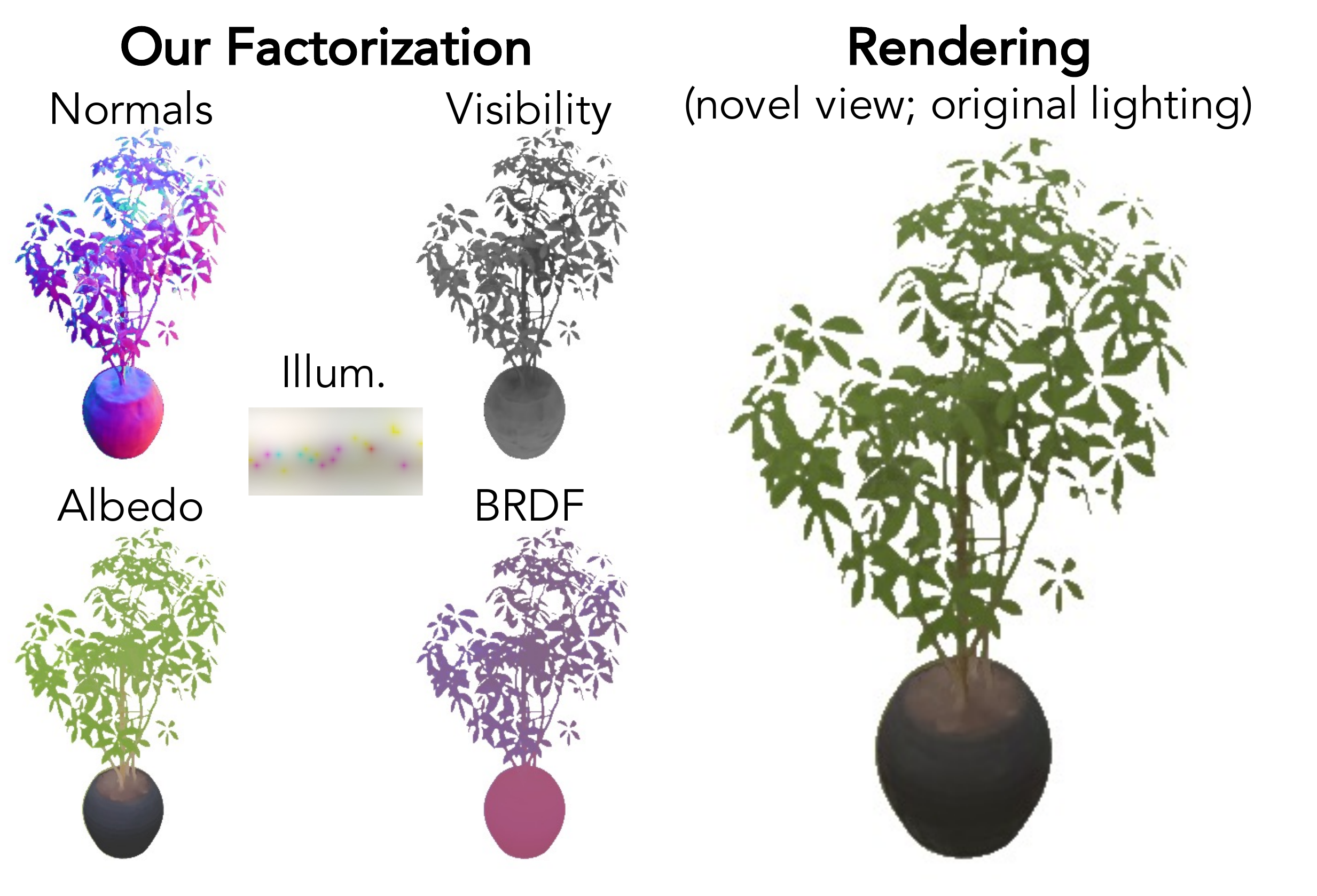}
  \caption{
  \textbf{Example factorization.}
  \model jointly solves for \emph{plausible} surface normals, light visibility, albedo, BRDFs, and lighting that together explain the observed views.
  Here we visualize light visibility as ambient occlusion and $z_\text{BRDF}$ directly as RGBs (similar colors indicate similar materials).
  }
  \label{fig:model_b}
\end{subfigure}

  \vspace{-1ex}
  \caption{
  \model is a coordinate-based model that factorizes, in an unsupervised manner, the appearance of a scene observed under one unknown lighting condition.
  It tackles this severely ill-posed problem by using a reconstruction loss, simple smoothness regularization, and data-driven BRDF priors.
  Modeling visibility explicitly, \model is a physically-based model that supports shadows under arbitrary lighting. 
  }
  \label{fig:model}
\end{figure*}

\rev{
We visualize the \model model and an example factorization it produces in \fig{fig:model}.
For implementation details including the network architecture, training paradigm, runtime, \etc, see \sect{sec:impl} of the appendix and \repo.
}

\subsection{Shape}

The input to our model is the same as what is used by NeRF \citep{mildenhall2020nerf}, so we can apply NeRF to our input images to compute initial geometry \rev{(though using Multi-View Stereo [MVS] geometry as initialization also works, as demonstrated in \sect{sec:mvs})}.
NeRF optimizes a neural radiance field: an MLP that maps from any 3D spatial coordinate and 2D viewing direction to the volume density at that 3D location and color emitted by particles at that location along the 2D viewing direction. 
\model leverages NeRF's estimated geometry by ``distilling'' it into a continuous surface representation that we use to initialize \model's geometry.
In particular, we use the optimized NeRF to compute the expected surface location along any camera ray, the surface normal at each point on the object's surface, and the visibility of light arriving from any direction at each point on the object's surface.
This subsection describes how we derive these functions from an optimized NeRF and how we re-parameterize them with MLPs so that they can be fine-tuned after this initialization step to improve the full re-rendering loss (\fig{fig:shape-results}).

\paragraph{Surface points}
Given a camera and a trained NeRF, we compute the location at which a ray $\bm{r}(t)=\bm{o}+t\bm{d}$ from that camera $\bm{o}$ along direction $\bm{d}$ is expected to terminate according to NeRF's optimized volume density $\sigma$:
\begin{equation}
\bm{x_\text{surf}}=\bm{o}+\left(\int_0^\infty T(t)\sigma\big(\bm{r}(t)\big)t\,dt\right)\bm{d}\,,
\end{equation}
where $T(t)=\exp\left(-\int_0^t\sigma\big(\bm{r}(s)\big)\,ds\right)$ is the probability that the ray travels distance $t$ without being blocked.
Instead of maintaining a full volumetric representation, we fix the geometry to lie on this surface distilled from the optimized NeRF.
This enables much more efficient relighting during both training and inference because we can compute the outgoing radiance just at each camera ray's expected termination location instead of every point along each camera ray.

\paragraph{Surface normals}
We compute analytic surface normals $\bm{n_\text{a}}(\bm{x})$ at any 3D location as the negative normalized gradient of NeRF's $\sigma$-volume \wrt $\bm{x}$.
Unfortunately, the normals derived from a trained NeRF tend to be noisy (\fig{fig:shape-results}) and therefore produce ``bumpy'' artifacts when used for rendering (see \suppvideo).
Therefore, we re-parameterize these normals using an MLP $\bm{f_\text{n}}$, which maps from any location $\bm{x_\text{surf}}$ on the surface to a ``denoised'' surface normal $\bm{n}$:
$\bm{f_\text{n}}: \bm{x_\text{surf}} \mapsto \bm{n}$.
During the joint optimization of \model's weights, we encourage the output of this MLP I) to stay close to the normals produced from the pretrained NeRF, II) to vary smoothly in the 3D space, and III) to reproduce the observed appearance of the object.
Specifically, the loss function reflecting I) and II)
is:
\begin{align}
\ell_\text{n}=\sum_{\bm{x_\text{surf}}}\bigg(&\frac{\lambda_1}{3}\big\Vert \bm{f_\text{n}}(\bm{x_\text{surf}}) - \bm{n_\text{a}}(\bm{x_\text{surf}})\big\Vert_2^2\\
+&\frac{\lambda_2}{3}\big\Vert \bm{f_\text{n}}(\bm{x_\text{surf}}) - \bm{f_\text{n}}(\bm{x_\text{surf}}+\bm{\epsilon}) \big\Vert_1\bigg)\,,
\end{align}
where $\bm{\epsilon}$ is a random 3D displacement from $\bm{x_\text{surf}}$ sampled from a zero-mean Gaussian with standard deviation $0.01$ ($0.001$ or $0.25$ for the real scenes due to different scene scales), and $\lambda_1$ and $\lambda_2$ are hyperparameters set to $0.1$ and $0.05$, respectively.
A similar smoothness loss on surface normals is used in the concurrent work by \citet{oechsle2021unisurf} for the goal of shape reconstruction.
Crucially, not restricting $\bm{x}$ on the expected surface increases the robustness of the MLP by providing a ``safe margin'' where the output remains well-behaved even when the input is slightly displaced from the surface. 
As shown in \fig{fig:shape-results}, \model's normal MLP produces normals that are significantly higher-quality than those produced by NeRF and are smooth enough to be used for relighting (\fig{fig:relight}).

\paragraph{Light visibility}
We compute the visibility $v_\text{a}$ to each light source from any point by marching through NeRF's $\sigma$-volume from the point to each light location, as in \citet{bi2020neural}.
However, similar to the estimated surface normals described above, the visibility estimates derived directly from NeRF's $\sigma$-volume are too noisy to be used directly (\fig{fig:shape-results}) and result in rendering artifacts (see \suppvideo).
We address this by re-parameterizing the visibility function as another MLP that maps from a surface location $\bm{x_\text{surf}}$ and a light direction $\bm{\omega_\text{i}}$ to the light visibility $v$:
$f_\text{v}: \left(\bm{x_\text{surf}}, \bm{\omega_\text{i}}\right)\mapsto v$.
We optimize the weights of $f_\text{v}$ to encourage the recovered visibility field I) to be close to the visibility traced from the NeRF, II) to be spatially smooth, and III) to reproduce the observed appearance.
Specifically, the loss function implementing I) and II) is:
\begin{align}
\ell_\text{v}=\sum_{\bm{x_\text{surf}}}\sum_{\bm{\omega_\text{i}}}\Big(&\lambda_3\big(f_\text{v}(\bm{x_\text{surf}}, \bm{\omega_\text{i}})-v_\text{a}(\bm{x_\text{surf}}, \bm{\omega_\text{i}})\big)^2\\
+&\lambda_4\big\vert f_\text{v}(\bm{x_\text{surf}}, \bm{\omega_\text{i}}) - f_\text{v}(\bm{x_\text{surf}}+\bm{\epsilon}, \bm{\omega_\text{i}})\big\vert\Big)\,,
\end{align}
where $\bm{\epsilon}$ is the random displacement defined above, and $\lambda_3$ and $\lambda_4$ are hyperparameters set to $0.1$ and $0.05$, respectively.
As the equation shows, smoothness is encouraged across spatial locations given the same $\bm{\omega_\text{i}}$, not the other way around.
This is by design, to avoid the visibility at a certain location getting blurred over different light locations.
Note that this is similar to the visibility fields in \citet{nerv} but in our case, we optimize the visibility MLP parameters to denoise the visibility derived from a pretrained NeRF and minimize the re-rendering loss.
For computing the NeRF visibility, we use a fixed set of $512$ light locations given a predefined illumination resolution (to be discussed later).
After optimization, $f_\text{v}$ produces spatially smooth and realistic estimates of light visibility, as can be seen in \fig{fig:shape-results} (II) and \fig{fig:joint-opt} (C), where we visualize the average visibility over all light directions (\ie, ambient occlusion).

In practice, before the full optimization of our model, we independently pretrain the visibility and normal MLPs to just reproduce the visibility and normal values from the NeRF $\sigma$-volume without any smoothness regularization or re-rendering loss.
This provides a reasonable initialization of the visibility maps, which prevents the albedo or Bidirectional Reflectance Distribution Function (BRDF) MLP from mistakenly attempting to explain away shadows as being modeled as ``painted on'' reflectance variation (see ``w/o geom.\ pretrain.'' in \tbl{tbl:quant} and \fig{fig:ablation}).

\subsection{Reflectance}

Our full BRDF model $\bm{R}$ consists of a diffuse component (Lambertian) fully determined by albedo $\bm{a}$ and a specular spatially-varying BRDF $\bm{f_\text{r}}$ (defined for any location on the surface $\bm{x_\text{surf}}$ with incoming light direction $\bm{\omega_\text{i}}$ and outgoing direction $\bm{\omega_\text{o}}$) learned from real-world reflectance:
\begin{equation}
\bm{R}(\bm{x_\text{surf}}, \bm{\omega_\text{i}}, \bm{\omega_\text{o}}) = \frac{\bm{a}(\bm{x_\text{surf}})}{\pi}+\bm{f_\text{r}}\left(\bm{x_\text{surf}}, \bm{\omega_\text{i}}, \bm{\omega_\text{o}}\right).
\end{equation}
Prior art in neural rendering has explored the use of parameterizing $\bm{f_\text{r}}$ with analytic BRDFs such as microfacet models \cite{bi2020neural,nerv} within a NeRF-like setting.
\rev{We also explore this ``analytic BRDF'' version of \model in \sect{sec:ablation}.}
Although these analytic models provide an effective BRDF parameterization for the optimization to explore, no prior is imposed upon the parameters themselves:
All materials that are expressible within a microfacet model are considered equally likely a priori.
Additionally, the use of an explicit analytic model limits the set of materials that can be recovered, and this may be insufficient for modeling all real-world BRDFs.

Instead of assuming an analytic BRDF, \model starts with a learned reflectance function that is pretrained to reproduce a wide range of empirically observed real-world BRDFs while also learning a latent space for those real-world BRDFs.
By doing so, we learn data-driven priors on real-world BRDFs that encourage the optimization to recover \emph{plausible} reflectance functions.
The use of such priors is crucial:
Because all of our observed images are taken under one (unknown) illumination, our problem is highly ill-posed, so priors are necessary to disambiguate the most likely factorization of the scene from the set of all possible factorizations.

\paragraph{Albedo}
We parameterize the albedo $\bm{a}$ at any surface location $\bm{x_\text{surf}}$ as an MLP $\bm{f_\text{a}}: \bm{x_\text{surf}} \mapsto \bm{a}$.
Because there is no direct supervision on albedo, and our model is only able to observe one illumination condition, we rely on simple spatial smoothness priors (and light visibility) to disambiguate between, \eg, the ``white-painted surface containing a shadow'' case and the ``black-and-white-painted surface'' case.
In addition, the reconstruction loss of the observed views also drives the optimization of $\bm{f_\text{a}}$.
The loss function that reflects this smoothness prior is:
\begin{equation}
\ell_\text{a}=\lambda_5\sum_{\bm{x_\text{surf}}}\frac{1}{3}\big\Vert \bm{f_\text{a}}(\bm{x_\text{surf}}) - \bm{f_\text{a}}(\bm{x_\text{surf}}+\bm{\epsilon}) \big\Vert_1\,,
\end{equation}
where $\bm{\epsilon}$ is the same random 3D perturbation as defined above, and $\lambda_5$ is a hyperparameter set to $0.05$.
The output from $\bm{f_\text{a}}$ is used as albedo in the Lambertian reflectance but not in the non-diffuse component, for which we assume the specular highlight color to be white. 
We empirically constrain the albedo prediction to $[0.03, 0.8]$ following \citet{ward1998rendering}, by scaling the network's final sigmoid output by $0.77$ and then adding a bias of $0.03$.
 
\paragraph{Learning priors from real-world BRDFs}
For the specular components of the BRDF, we seek to learn a latent space of real-world BRDFs and a paired ``decoder'' that translates each latent code in the learned space $\bm{z_\text{BRDF}}$ to a full 4D BRDF.
To this end, we adopt the Generative Latent Optimization (GLO) approach \cite{bojanowski2017optimizing}, which has been previously used by other coordinate-based models such as \citet{park2019deepsdf} and \citet{martin2020nerf}.
The $\bm{f_\text{r}}$ component of our model is pretrained using the the MERL dataset \cite{matusik2003}.
Because the MERL dataset assumes isotropic materials, we parameterize the incoming and outgoing directions for $\bm{f_\text{r}}$ using Rusinkiewicz coordinates \cite{rusinkiewicz1998new} $(\phi_\text{d}, \theta_\text{h}, \theta_\text{d})$ (3 degrees of freedom) instead of $\bm{\omega_\text{i}}$ and $\bm{\omega_\text{o}}$ (4 degrees of freedom).
Denote this coordinate conversion by
$\bm{g}: (\bm{n}, \bm{\omega_\text{i}}, \bm{\omega_\text{o}})\mapsto (\phi_\text{d}, \theta_\text{h}, \theta_\text{d})$,
where $\bm{n}$ is the surface normal at that point.
We train a function $\bm{f_\text{r}}'$ (a re-parameterization of $\bm{f_\text{r}}$) that maps from a concatenation of a latent code $\bm{z_\text{BRDF}}$ (which represents a BRDF identity) and Rusinkiewicz coordinates $(\phi_\text{d}, \theta_\text{h}, \theta_\text{d})$ to an achromatic reflectance $\bm{r}$:
\begin{equation}
\bm{f_\text{r}}': \big(\bm{z_\text{BRDF}}, (\phi_\text{d}, \theta_\text{h}, \theta_\text{d})\big) \mapsto \bm{r}\,.
\end{equation}
To train this model, we optimize both the weights of the MLP and the set of latent codes $\bm{z_\text{BRDF}}$ to reproduce a set of real-world BRDFs.
Simple mean squared errors are computed on the log of the High-Dynamic-Range (HDR) reflectance values to train $\bm{f_\text{r}}'$.

Because the color component of our reflectance model is assumed to be handled by the albedo MLP, we discard all color information from the MERL dataset by converting its RGB reflectance values into achromatic ones%
\footnote{
In principle, one should perform diffuse-specular separation on the MERL BRDFs and then learn priors on just the specular lobes.
We experimented with this idea by using the separation provided by \citet{sun2018connecting}, but this yielded qualitatively worse results.}.
The latent BRDF identity codes $\bm{z_\text{BRDF}}$ are parameterized as unconstrained 3D vectors and initialized with a zero-mean isotropic Gaussian with a standard deviation of $0.01$.
No sparsity or norm penalty is imposed on $\bm{z_\text{BRDF}}$ during training. 
After this pretraining, the weights of this BRDF MLP are frozen during the joint optimization of our entire model, and we predict only $\bm{z_\text{BRDF}}$ for each $\bm{x_\text{surf}}$ by training from scratch a BRDF identity MLP (\fig{fig:model_a}): $\bm{f_\text{z}}: \bm{x_\text{surf}}\mapsto \bm{z_\text{BRDF}}$.
This can be thought of as predicting spatially-varying BRDFs for all the surface points in the plausible space of real-world BRDFs.
We optimize the BRDF identity MLP to minimize the re-rendering loss and the same spatial smoothness prior as in albedo:
\begin{equation}
\ell_\text{z}=\lambda_6\sum_{\bm{x_\text{surf}}}
\frac{\big\Vert \bm{f_\text{z}}(\bm{x_\text{surf}}) - \bm{f_\text{z}}(\bm{x_\text{surf}} + \bm{\epsilon}) \big\Vert_1}{\text{dim}(\bm{z_\text{BRDF}})}\,,
\end{equation}
where $\lambda_6$ is a hyperparameter set to $0.01$, and $\text{dim}(\bm{z_\text{BRDF}})$ denotes the dimensionality of the BRDF latent code ($3$ in our implementation because there are only $100$ materials in the MERL dataset).
The final BRDF is the sum of the Lambertian component and the learned non-diffuse reflectance (subscript of $\bm{x_\text{surf}}$ dropped for brevity):
\begin{equation}
\bm{R}(\bm{x},\bm{\omega_\text{i}},\bm{\omega_\text{o}})=\frac{\bm{f_\text{a}}(\bm{x})}{\pi}+\bm{f_\text{r}}'\Big(\bm{f_\text{z}}(\bm{x}),
\bm{g}\big(\bm{f_\text{n}}(\bm{x}), \bm{\omega_\text{i}}, \bm{\omega_\text{o}}\big)
\Big)\,,
\end{equation}
where the specular highlight color is assumed to be white.

\subsection{Lighting}

We adopt a simple and direct representation of lighting: an HDR light probe image \cite{debevec1998rendering} in the latitude-longitude format.
In contrast to spherical harmonics or a mixture of spherical Gaussians, this representation allows our model to represent detailed high-frequency lighting and therefore to support hard cast shadows.
That said, the challenges of using this representation are clear:
It contains a large number of parameters, and every pixel/parameter can vary independently of all other pixels.
This issue can be ameliorated by our use of the light visibility MLP, which allows us to quickly evaluate a surface point's visibility to all pixels of the light probe.
Empirically, we use a $16\times32$ resolution for our lighting environments, as we do not expect to recover higher-frequency content beyond that resolution (lighting is effectively low-pass filtered by the object's BRDFs \cite{ramamoorthi01}, and our objects are not shiny or mirror-like).

To encourage smoother lighting, we apply a simple $\ell^2$ gradient penalty on the pixels of the light probe $\bm{L}$ along both the horizontal and vertical directions:
\begin{equation}
\ell_\text{i}=\lambda_7\left( \Big\Vert \begin{bmatrix}-1 & 1\end{bmatrix} * \bm{L} \Big\Vert_2^2 + \Bigg\Vert \begin{bmatrix}-1 \\ 1\end{bmatrix} * \bm{L} \Bigg\Vert_2^2 \right)\,,
\end{equation}
where $*$ denotes the convolution operator, and $\lambda_7$ is a hyperparameter set to $5\times 10^{-6}$ (given that there are $512$ pixels with HDR values).
During the joint optimization, these probe pixels get updated directly by the final reconstruction loss and the gradient penalty.

\subsection{Rendering}

Given the surface normal, visibility for all light directions, albedo, and BRDF at each point $\bm{x_\text{surf}}$, as well as the estimated lighting, the final physically-based, non-learnable renderer renders an image that is then compared against the observed image.
The errors in this rendered image are backpropagated up to, but excluding, the $\sigma$-volume of the pretrained NeRF, thereby driving the joint estimation of surface normals, light visibility, albedo, BRDFs, and lighting.

Given the ill-posed nature of the problem (largely due to our only observing one unknown illumination), we expect the majority of useful information to be from direct illumination rather than global illumination and therefore consider only single-bounce direct illumination (\ie, from the light source to the object surface then to the camera).
This assumption also reduces the computational cost of evaluating our model.
Mathematically, the rendering equation in our setup is (subscript of $\bm{x_\text{surf}}$ dropped again for brevity):
\begin{align}
\bm{L_\text{o}}(\bm{x}, \bm{\omega_\text{o}}) = \int_\Omega \bm{R}(\bm{x}, \bm{\omega_\text{i}}, \bm{\omega_\text{o}}) \bm{L_\text{i}}(\bm{x}, \bm{\omega_\text{i}}) \big(\bm{\omega_\text{i}}\cdot \bm{n}(\bm{x})\big) d\bm{\omega_\text{i}}\\
= \sum_{\bm{\omega_\text{i}}} \bm{R}(\bm{x}, \bm{\omega_\text{i}}, \bm{\omega_\text{o}}) \bm{L_\text{i}}(\bm{x}, \bm{\omega_\text{i}}) \big(\bm{\omega_\text{i}}\cdot \bm{f_\text{n}}(\bm{x})\big) \Delta\bm{\omega_\text{i}} = \sum_{\bm{\omega_\text{i}}} \bigg(\frac{\bm{f_\text{a}}(\bm{x})}{\pi} + \\
\bm{f_\text{r}}'\Big(\bm{f_\text{z}}(\bm{x}), \bm{g}\big(\bm{f_\text{n}}(\bm{x}), \bm{\omega_\text{i}}, \bm{\omega_\text{o}}\big)\Big) \bigg) \bm{L_\text{i}}(\bm{x}, \bm{\omega_\text{i}}) \big(\bm{\omega_\text{i}}\cdot \bm{f_\text{n}}(\bm{x})\big) \Delta\bm{\omega_\text{i}}\,,
\end{align}
where $\bm{L_\text{o}}(\bm{x}, \bm{\omega_\text{o}})$ is the outgoing radiance at $\bm{x}$ as viewed from $\bm{\omega_\text{o}}$, $\bm{L_\text{i}}(\bm{x}, \bm{\omega_\text{i}})$ is the incoming radiance, masked by the visibility $f_\text{v}(\bm{x}, \bm{\omega_\text{i}})$, arriving at $\bm{x}$ along $\bm{\omega_\text{i}}$ directly from a light probe pixel (since we consider only single-bounce direct illumination), and $\Delta\bm{\omega_\text{i}}$ is the solid angle corresponding to the lighting sample at $\bm{\omega_\text{i}}$.

The final reconstruction loss $\ell_\text{recon}$ is simply the mean squared error (with a unit weight) between the rendering and the observed image.
Therefore, our full loss function is the summation of all the previously defined losses: $\ell_\text{recon} + \ell_\text{n} + \ell_\text{v} + \ell_\text{a} + \ell_\text{z} + \ell_\text{i}$. %
\section{Results \& Applications}

In this section, we show
I) the high-quality geometry achieved by \model,
II) \model's capability of jointly estimating shape, reflectance, and lighting,
III) the application of free-viewpoint relighting, with a single point light or any arbitrary light probe (\fig{fig:relight} and \fig{fig:real}), enabled by this capability,
IV) \model's performance when using MVS instead of NeRF for shape initialization, and finally
V) the application of material editing (\fig{fig:mat-edit}).

\rev{
See \sect{sec:data} of the appendix for how the various types of data used in this work are rendered, captured, or collected.
}

\subsection{Shape Optimization}

\begin{figure*}[!htbp]
  \centering
  \includegraphics[width=\textwidth]{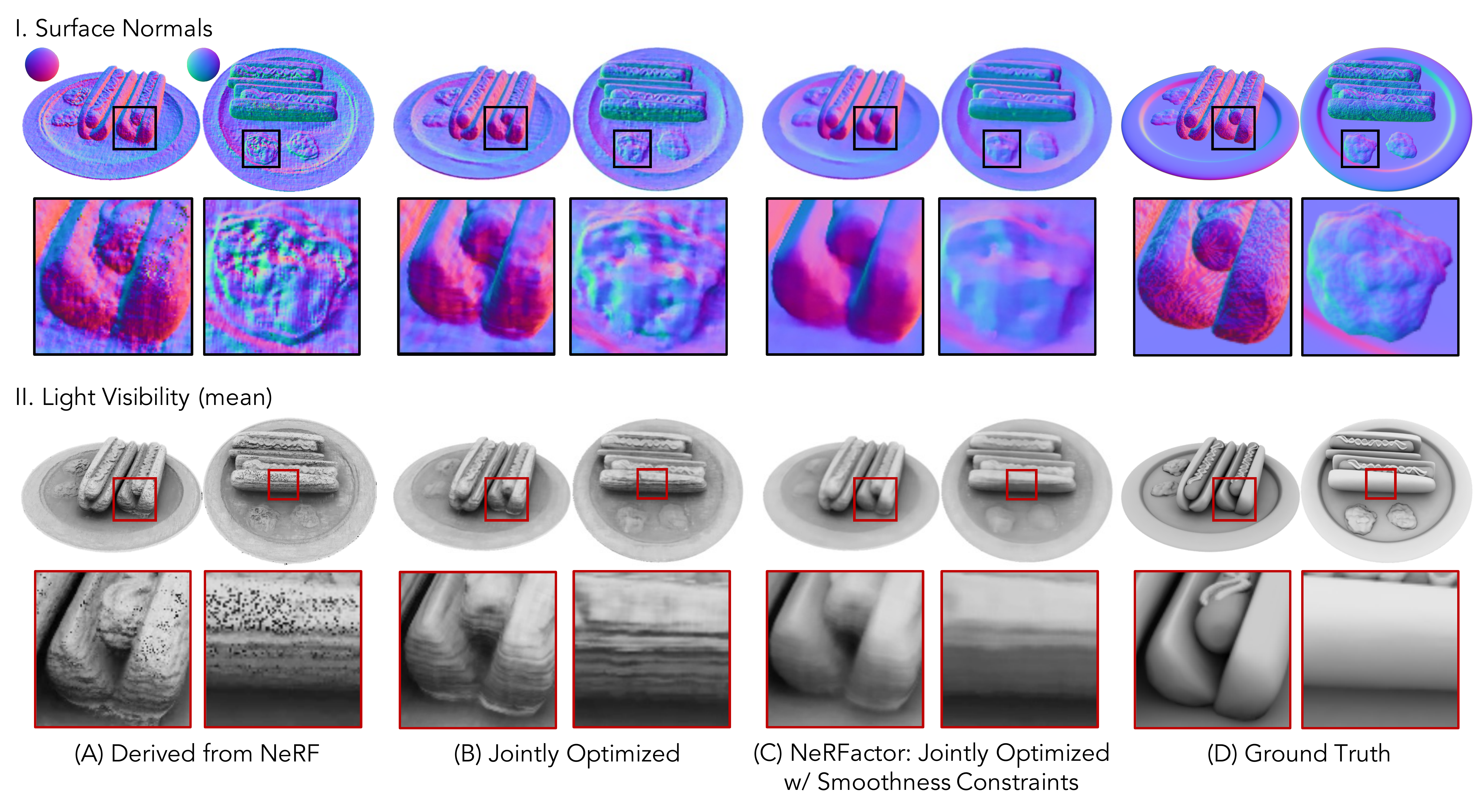}
  \vspace{-6ex}
  \caption{
  \textbf{High-quality geometry recovered by \model.}
  (A) We can directly derive the surface normals and light visibility from a trained NeRF.
  However, geometry derived in this way is too noisy to be used for relighting (see \suppvideo).
  (B) Jointly optimizing shape and reflectance improves the NeRF geometry, but there is still significant noise (\eg, the stripe artifacts in II).
  (C) Joint optimization with smoothness constraints leads to smooth surface normals and light visibility that resemble ground truth.
  Visibility averaged over all incoming light directions is ambient occlusion.
  }
  \label{fig:shape-results}
\end{figure*}

\model jointly estimates an object's shape in the form of surface points and their associated surface normals as well as their visibility to each light location.
\fig{fig:shape-results} visualizes these geometric properties.
To visualize light visibility, we take the per-pixel mean of the $512$ visibility maps corresponding to each pixel of a $16\times32$ light probe, and visualize that average map (\ie, ambient occlusion) as a grayscale image.
See \suppvideo for movies of per-light visibility maps (\ie, shadow maps).
As \fig{fig:shape-results} shows, our surface normals and light visibility are smooth and resemble the ground truth, thanks to the joint estimation procedure that minimizes re-rendering errors and encourages spatial smoothness.

If we ablate the spatial smoothness constraints and rely on only the re-rendering loss, we end up with noisy geometry that is insufficient for rendering.
Although these geometry-induced artifacts may not show up under low-frequency lighting, harsh lighting conditions (such as a single point light with no ambient illumination, \ie, One-Light-at-A-Time [OLAT]) reveal them as demonstrated in \suppvideo.
Perhaps surprisingly, even when our smoothness constraints are disabled, the geometry estimated by \model is still significantly less noisy than the original NeRF geometry (compare [A] with [B] of \fig{fig:shape-results} and see [I] of \tbl{tbl:quant}) because the re-rendering loss encourages smoother geometry.
See \sect{sec:ablation} for more details.

\subsection{Joint Estimation of Shape, Reflectance, \& Lighting}
\label{sec:joint}
\dummytext

In this experiment, we demonstrate how \model factorizes appearance into shape, reflectance, and illumination for scenes with complex geometry and/or reflectance. 

When visualizing albedo, we adopt the convention used by the intrinsic image literature of assuming that the absolute brightness of albedo and shading is unrecoverable \cite{Land71lightnessand}, and furthermore we assume that the problem of color constancy (solving for a global color correction that disambiguates between the average color of the illuminant and the average color of the albedo \cite{Buchsbaum80}) is also out of scope.
In accordance with these two assumptions, we visualize our predicted albedo and measure its accuracy by first scaling each RGB channel by a global scalar that is identified so as to minimize the mean squared error \wrt the ground-truth albedo%
\footnote{As such, such corrections are impossible for real scenes where the ground-truth albedo is unavailable.},
as is done by \citet{Barron2015Shape}.
Unless stated otherwise, all albedo predictions for the synthetic scenes are corrected this way, and we apply gamma correction ($\gamma=2.2$) to display them properly in the figures.
Our estimated light probes are not scaled this way \wrt the ground truth (since lighting estimation is not the primary goal of this work) and are visualized by simply scaling their maximum intensity across all RGB channels to $1$ and then applying gamma correction ($\gamma=2.2$).

\begin{figure*}[!htbp]
  \centering
  \includegraphics[width=\textwidth]{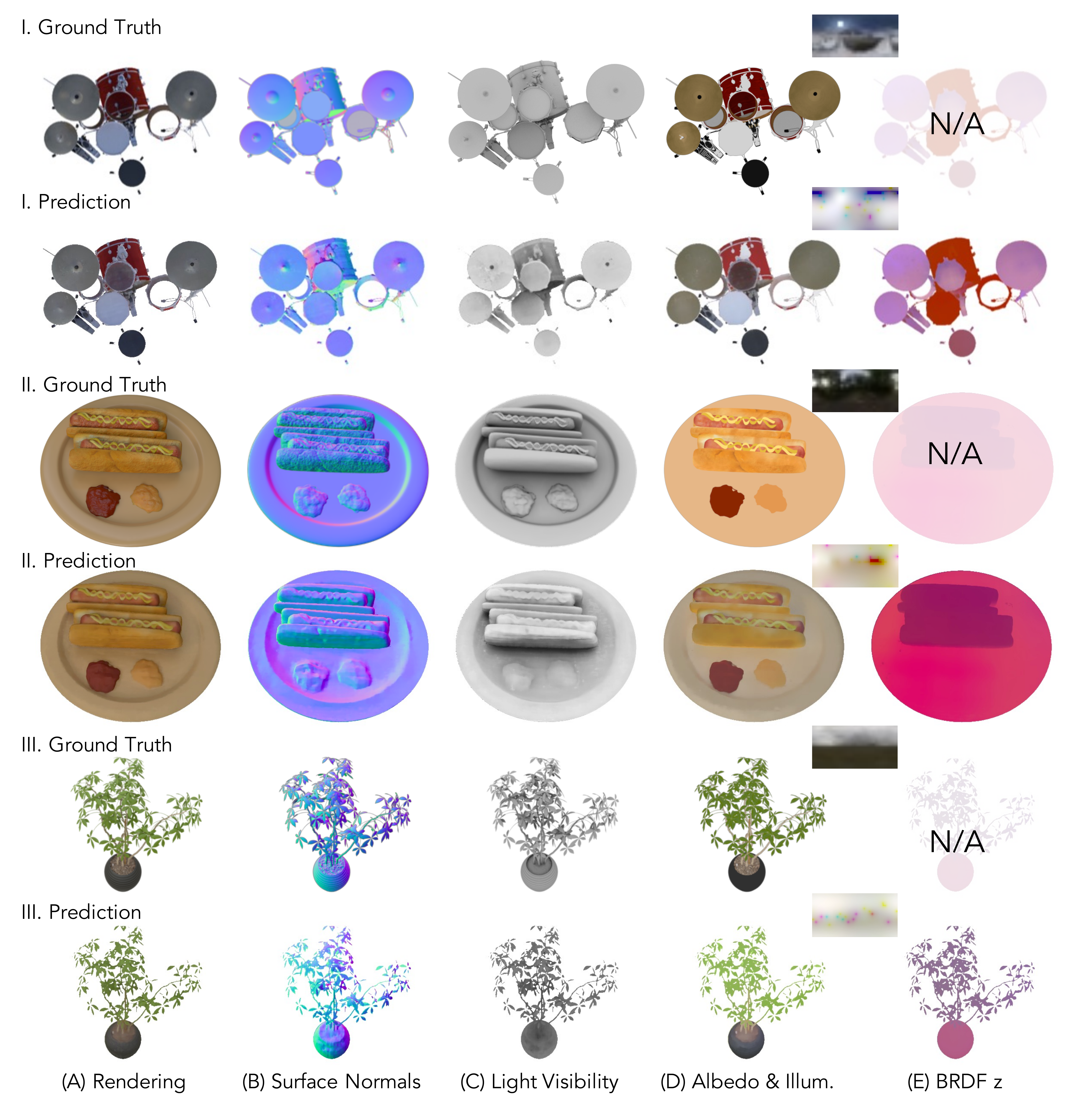}
  \vspace{-6ex}
  \caption{
  \textbf{Joint optimization of shape, reflectance, and lighting.}
  Although our recovered surface normals, visibility, and albedo sometimes omit some fine-grained detail, they still closely resemble the ground truth.
  Despite that lighting recovered by \model is heavily oversmoothed (because our objects are not shiny)
  and incorrect on the bottom half of the hemisphere (since objects are only ever observed from the top hemisphere), the dominant light sources and occluders are localized nearby their ground-truth locations in the light probes.
  Note that we are unable to compare against ground-truth BRDFs, as they are defined using Blender's shader node trees, while our recovered BRDFs are parameterized by our learned model. 
  }
  \label{fig:joint-opt}
\end{figure*}

As shown in \fig{fig:joint-opt} (B), \model predicts high-quality and smooth surface normals that are close to the ground truth except in regions with very high-frequency details such as the bumpy surface of the hot dog buns.
In \texttt{drums}, we see that \model successfully reconstructs fine details such as the screw at the cymbal center and the metal rims on sides of the drums.
For \texttt{ficus}, \model recovers the complex leaf geometry.
The ambient occlusion maps also correctly portray the average exposure of each point in the scene to the lights.
Albedo is recovered cleanly with barely any shadowing or shading detail inaccurately attributed to albedo variation; note how the shading on the drums is absent in the albedo prediction.
Moreover, the predicted light probes correctly reflect the locations of the primary light sources and the blue sky (blue pixels in [I]).
In all three scenes, the predicted BRDFs are spatially-varying and correctly reflect that different parts of the scene have different materials, as indicated by different BRDF latent codes in (E). 

\begin{figure*}[!htbp]
  \centering
  \includegraphics[width=0.88\textwidth]{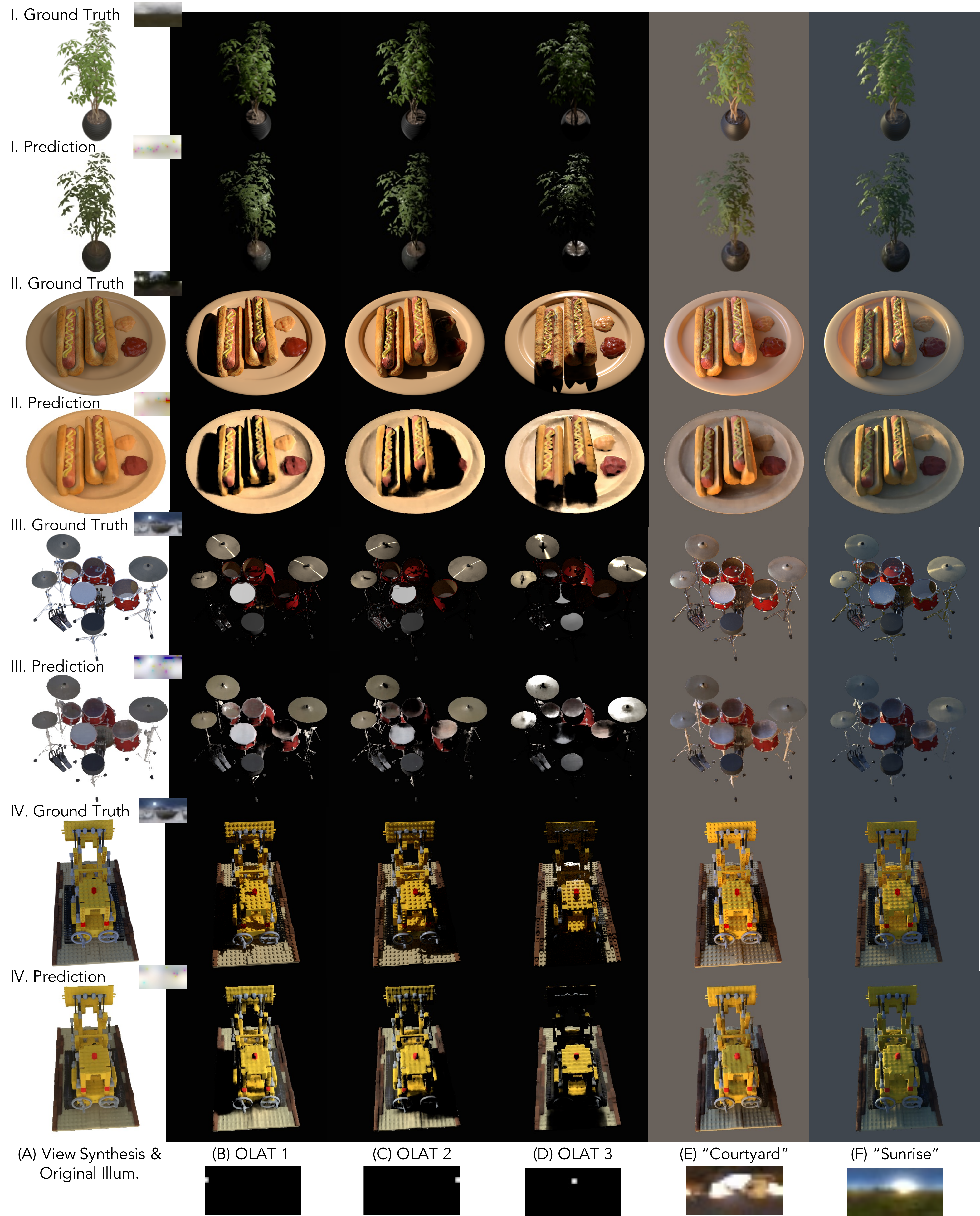}
  \caption{
  \textbf{Free-viewpoint relighting.}
  The factorization that \model produces can be used to perform ``free-viewpoint relighting'': rendering a novel view of the object under arbitrary lighting conditions including the challenging OLAT conditions.
  The renderings produced by \model qualitatively resemble the ground truth and accurately exhibit challenging effects such as specularities and cast shadows (both hard and soft).
  }
  \label{fig:relight}
\end{figure*} 
\begin{figure*}[!htbp]
  \centering
  \includegraphics[width=\textwidth]{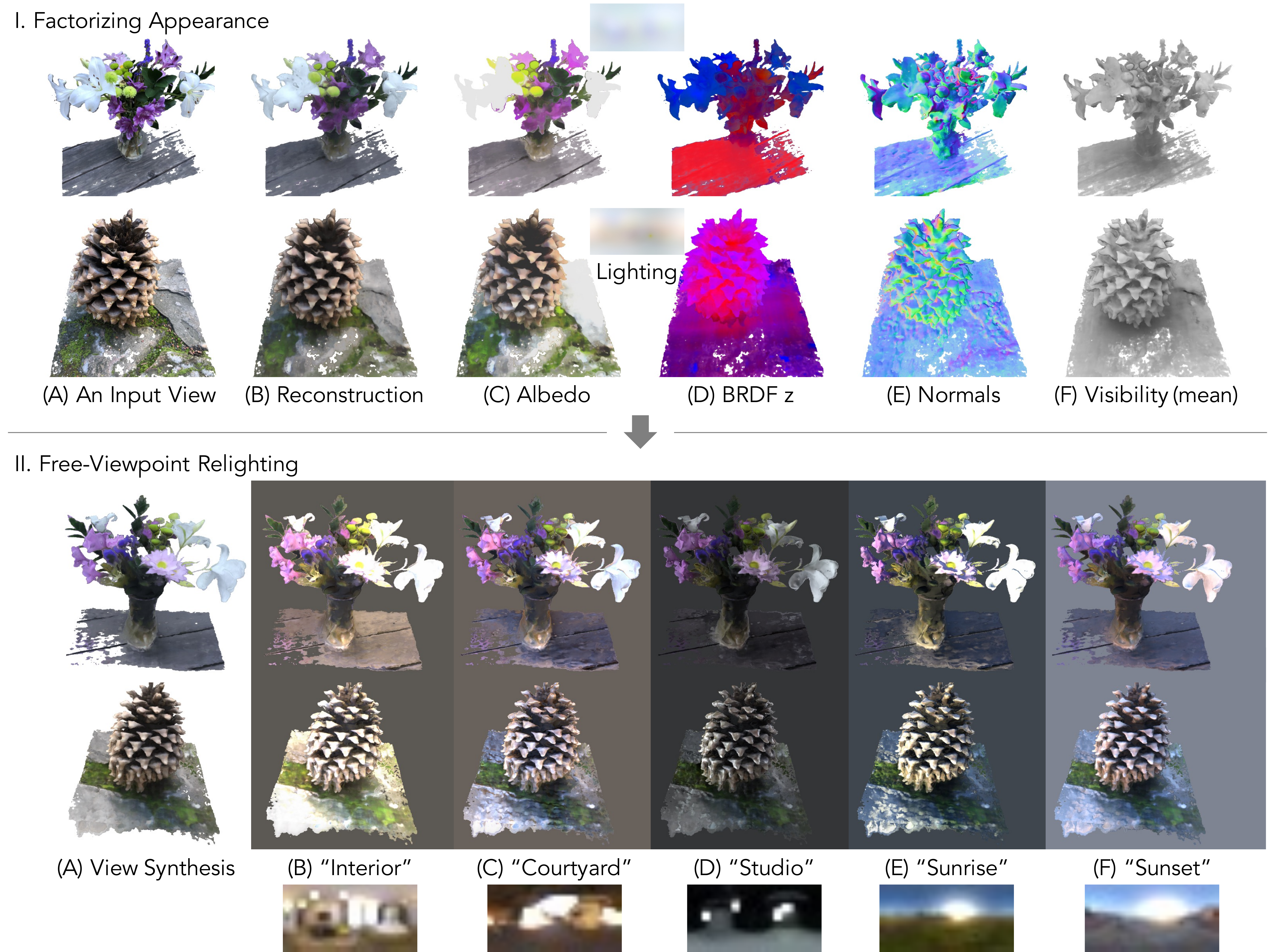}
  \vspace{-4ex}
  \caption{
  \textbf{Results of real-world captures.} 
  (I) Given images of a real-world object lit by unknown lighting (A), \model factorizes its appearance into albedo (C), spatially-varying BRDF latent codes (D), surface normals (E), and light visibility for all incoming light directions (visualized here as ambient occlusion; F).
  Note how the estimated flower albedo is shading-free. 
  (II) With this factorization, one can synthesize novel views of the scene relit by any arbitrary lighting.
  Even on these challenging real-world scenes, \model is able to synthesize realistic specularities and shadows across various lighting conditions.
  }
  \label{fig:real}
\end{figure*} 
\begin{figure*}[!htbp]
  \centering
  \includegraphics[width=\textwidth]{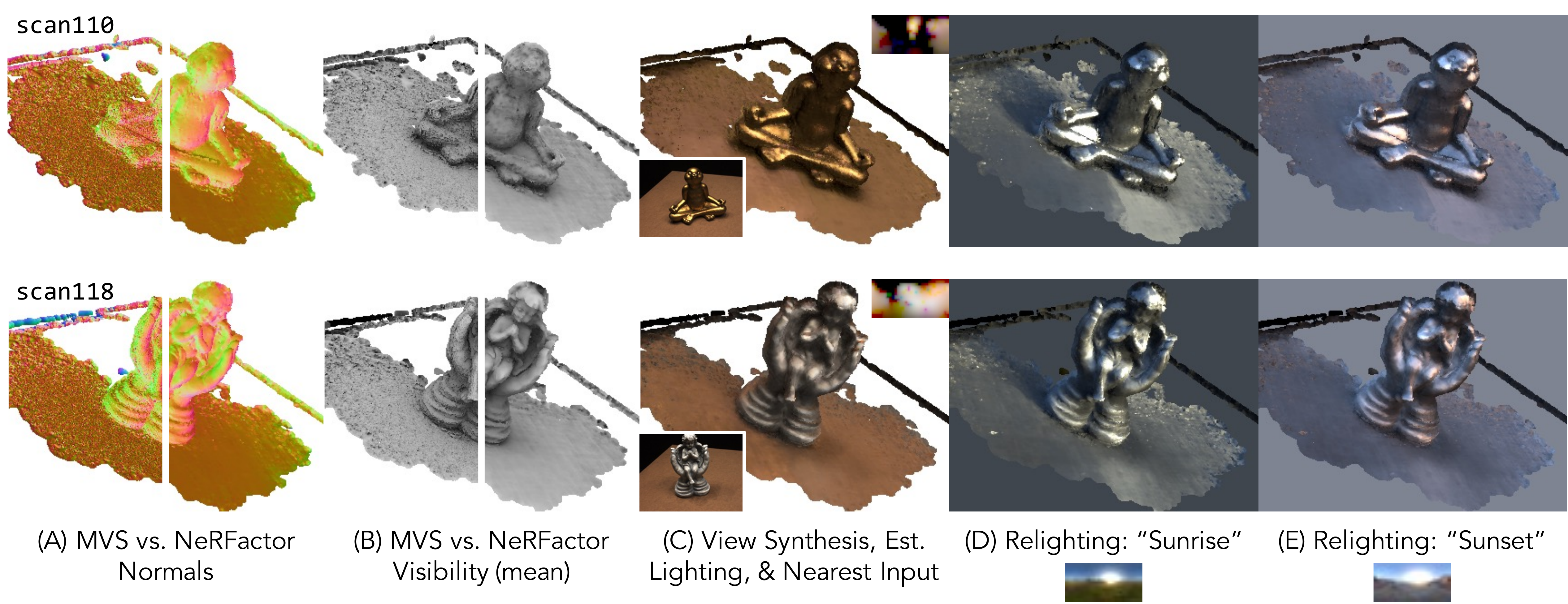}
  \vspace{-5ex}
  \caption{
  \rev{
  \textbf{Results of real-world captures when using MVS for shape initialization.}
  (A, B) We demonstrate how \model smooths out the noisy MVS geometry while preserving its details.
  (C, D, E) With higher-quality geometry, we can perform realistic view synthesis and relighting (note the shadows in [D]).
  See \sect{sec:mvs} for more discussions.
  }
  }
  \label{fig:mvs}
\end{figure*} 
Instead of representing lighting with a more sophisticated representation such as spherical harmonics, we opt for a straightforward representation: a latitude-longitude map whose pixels are HDR intensities.
Because lighting is effectively convolved by a low-pass filter when reflected by a moderately diffuse BRDF \cite{ramamoorthi01}, we do not expect to recover lighting at a resolution higher than $16\times32$.
As shown in \fig{fig:joint-opt} (I), \model estimates a light probe that correctly captures the bright light source on the far left and the blue sky.
Similarly, in \fig{fig:joint-opt} (II), the dominant light source location is also correctly estimated (the bright white blob on the left).

\subsection{Free-Viewpoint Relighting}

\model estimates 3D fields of shape and reflectance, thus enabling simultaneous relighting and view synthesis.
As such, all the relighting results shown in this paper and \suppvideo are rendered from novel viewpoints.
To probe the limits of \model, we use harsh test lighting conditions that have one point light on at a time (OLAT), with no ambient illumination.
These test illuminations induce hard cast shadows, which effectively exposes rendering artifacts due to inaccurate geometry or materials.
For visualization purposes, we composite the relit results (using NeRF's predicted opacity or MVS' mesh silhouettes) onto backgrounds whose colors are the averages over upper halves of the light probes.

As shown in \fig{fig:relight} (II), \model synthesizes correct hard shadows cast by the hot dogs under the three test OLAT conditions.
\model also produces realistic renderings of the ficus under the OLAT conditions (I), especially when the ficus is back-lit by the point light in (D).
Note that the ground truth in (D) appears brighter than \model's results because \model models only direct illumination, whereas the ground-truth image was rendered with global illumination.
When we relight the objects with two new light probes, realistic soft shadows are synthesized on the \texttt{hotdog} plate (II).
In \texttt{ficus}, specularities on the vase correctly reflect the primary light sources in both test probes.
The leaves also exhibit realistic specular highlights close to the ground truth in (F).
In \texttt{drums} (III), the cymbals are correctly estimated to be specular and exhibit realistic reflection, though different from the ground-truth anisotropic reflection (D).
This is as expected because all MERL BRDFs are isotropic \cite{matusik2003}.
Though unable to explain these anisotropic reflections, \model correctly leaves them out in albedo rather than interprets them as albedo paints, since doing that would violate the albedo smoothness constraint and contradict those reflections' view dependency.
In \texttt{lego}, realistic hard shadows are synthesized by \model for the OLAT test conditions (IV).

\paragraph{Relighting real scenes}
We apply \model to the two real scenes, \texttt{vasedeck} and \texttt{pinecone}, captured by \citet{mildenhall2020nerf}.
These captures are particularly suitable for \model:
There are around $100$ multi-view images of each scene lit by an unknown environment lighting.
As in NeRF, we run COLMAP Structure From Motion (SFM) \cite{schoenberger2016sfm} to obtain the camera intrinsics and extrinsics for each view.
We then train a vanilla NeRF to obtain an initial shape estimate, which we distill into \model and jointly optimize together with reflectance and illumination.
As \fig{fig:real} (I) shows, the appearance is factorized into lighting and 3D fields of surface normals, light visibility, albedo, and spatially-varying BRDF latent codes that together explain the observed views.
With such factorization, we relight the scenes by replacing the estimated illumination with novel arbitrary light probes (\fig{fig:real} [II]).
Because our factorization is fully 3D, all the intermediate buffers can be rendered from any viewpoint, and the relighting results shown are also from novel viewpoints.
Note that bound these real scenes within 3D boxes to avoid faraway geometry blocking light from certain directions and casting shadows during relighting.

\subsection{Shape Initialization Using Multi-View Stereo}
\label{sec:mvs}

\rev{
We have demonstrated how \model uses the geometry extracted from NeRF as an initialization, and continues to refine this geometry while factorizing reflectance and lighting jointly.
Here we explore whether \model can work with other shape initializations such as MVS.
Specifically, we consider the DTU-MVS dataset \cite{jensen2014large,aanaes2016large} that provides around $50$ multi-view images (and their corresponding camera poses) for each scene.
We initialize \model's shape with the Poisson reconstruction \cite{Kazhdan2006Poisson} of the MVS reconstruction by \citet{furukawa2009accurate}.
See \sect{sec:data} of the appendix for more details on these data.
This experiment explores not only another possibility for shape initialization but also one more source of real images that \model is evaluated on.
}

\rev{
\model achieves high-quality shape estimation when starting from MVS geometry instead of NeRF geometry.
As \fig{fig:mvs} (A, B) demonstrates, the surface normals and light visibility estimated by \model are free of the noise MVS suffers from and meanwhile possess enough geometric details.
With these higher-quality geometry estimates, \model achieves realistic view synthesis results that resemble the nearest neighbor input images (\fig{fig:mvs} [C]).
The shiny material of \texttt{scan110} indeed facilitates the recovery of a higher-frequency lighting condition (compare the two lighting conditions recovered in [C]).
We then further relight the scenes, from this novel viewpoint, with two novel light probes, as shown in (D, E).
In addition to the realistic specular highlights, notice also the shadows synthesized by \model in (D), thanks to its visibility modeling.
Note that \model opts to explain \texttt{scan110} with white albedo and gold lighting (instead of the other way around) due to the fundamental ambiguity discussed in \sect{sec:joint}, but still manages to relight the scene realistically using this plausible explanation.
}

\subsection{Material Editing}

\begin{figure}[!htbp]
  \centering
  \includegraphics[width=\linewidth]{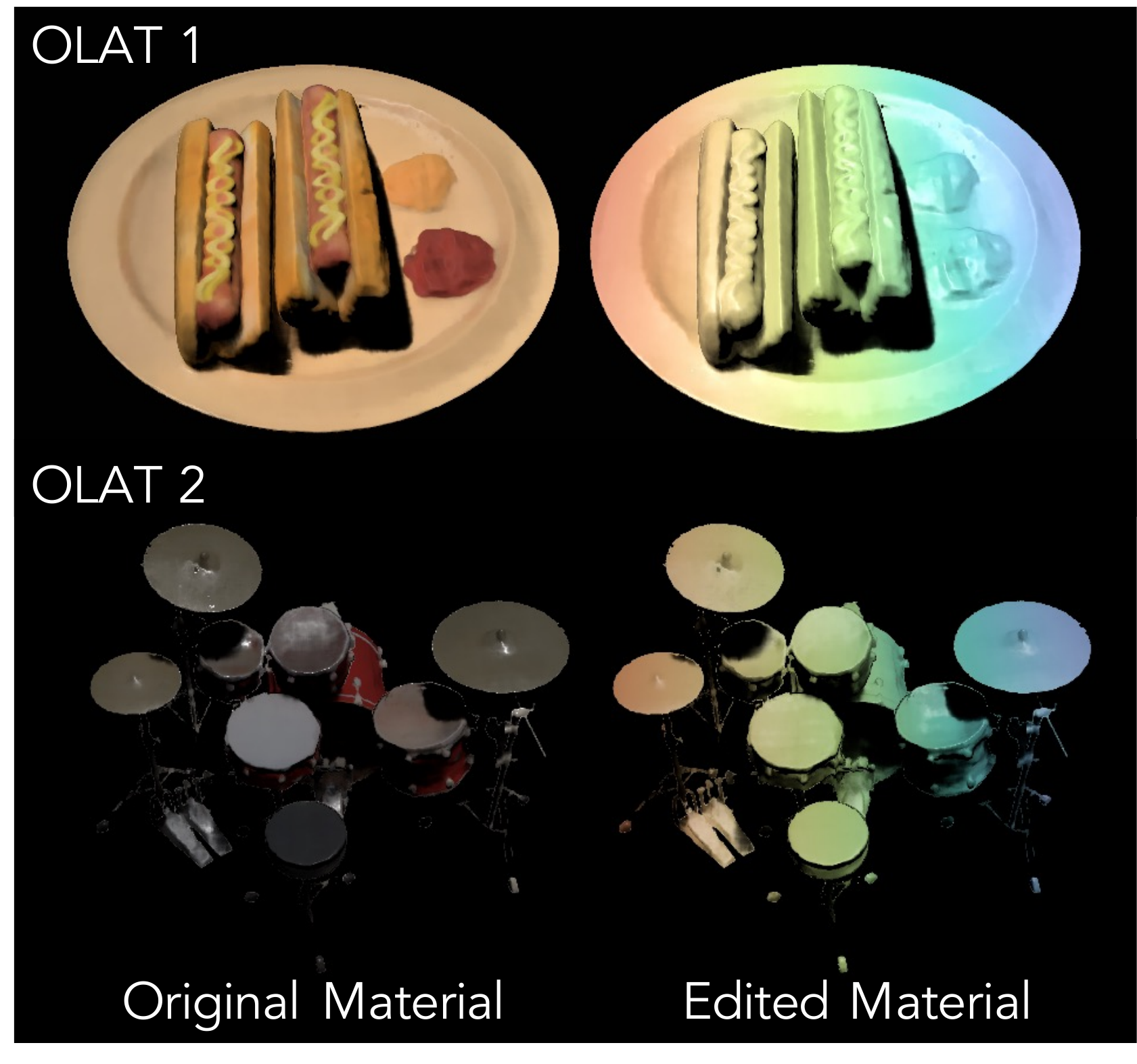}
  \vspace{-5ex}
  \caption{
  \textbf{Material editing and relighting.}
  With the \model factorization, we show the original materials relit by two OLAT conditions (Left) alongside the edited materials relit by the same OLAT conditions (Right).
  See the text for how we modified the albedo and reflectance.
  }
  \label{fig:mat-edit}
\end{figure}

Since \model factorizes diffuse albedo and specular BRDF from appearance, one can edit the albedo, non-diffuse BRDF, or both, and then re-render the edited object under an arbitrary lighting condition from any viewpoint.
Here we override the estimated $\bm{z_\text{BRDF}}$ to the learned latent code of \texttt{pearl-paint} in the MERL dataset and the estimated albedo to colors linearly interpolated from the \texttt{turbo} colormap, spatially varying based on the surface points' $x$-coordinates.
As \fig{fig:mat-edit} (Left) demonstrates, with the factorization by \model, we are able to realistically relight the original estimated materials with the two challenging OLAT conditions.
Furthermore, the edited materials are also relit with realistic specular highlights and hard shadows by the same test OLAT conditions (\fig{fig:mat-edit} [Right]).

\section{Evaluation Studies}

In this section, we perform ablation studies to evaluate the importance of each model component and compare \model against both classic and deep learning-based state of the art in the tasks of appearance factorization and relighting.
For quantitative evaluations, we use as metrics Peak Signal-to-Noise Ratio (PSNR), Structural Similarity Index Measure (SSIM) \cite{wang2004image}, and Learned Perceptual Image Patch Similarity (LPIPS) \cite{zhang_unreasonable_2018}.

\rev{
See also \sect{sec:consistency} of the appendix for whether albedo estimation for the same object remains consistent across different input lighting conditions.
}

\subsection{Ablation Studies}
\label{sec:ablation}
\dummytext

In this section, we compare \model against other reasonable design alternatives by ablating each of the major model components and observing whether there is performance drop quantitatively.

\rev{
We present the quantitative ablation studies in \tbl{tbl:quant}.
For the qualitative ablation studies, see \sect{sec:qualiablation} of the appendix and \suppvideo.
}

\begin{table*}[!htbp]
\centering
    \caption{\textbf{Quantitative evaluation.}
    Reported numbers are the arithmetic means of all four synthetic scenes (\texttt{hotdog}, \texttt{ficus}, \texttt{lego}, and \texttt{drums}) over eight uniformly sampled novel views.
    The top three performing techniques for each metric are highlighted in red, orange, and yellow, respectively.
    For Tasks IV and V, we relight the scenes with $16$ novel lighting conditions: eight OLAT conditions plus the eight light probes included in Blender.
    We are unable to present normal, view synthesis, or relighting metrics for SIRFS since it does not support non-orthographic cameras or ``world-space'' geometry (although \fig{fig:sirfs} shows that the geometry recovered by SIRFS is inaccurate).
    See \sect{sec:ablation} for discussion and \sect{sec:qualiablation} for qualitative ablation studies.
    }
    \label{tbl:quant}
    \vspace{-1ex}
    \resizebox{\textwidth}{!}{ 
    \begin{tabular}{lccccccccccccc}
        \toprule
         & \multicolumn{1}{c}{\textbf{I. Normals}} & \multicolumn{3}{c}{\textbf{II. Albedo}} & \multicolumn{3}{c}{\textbf{III. View Synthesis}} & \multicolumn{3}{c}{\textbf{IV. FV Relighting (point)}} & \multicolumn{3}{c}{\textbf{V. FV Relighting (image)}} \\
        \cmidrule(lr){2-2} \cmidrule(lr){3-5} \cmidrule(lr){6-8} \cmidrule(lr){9-11} \cmidrule(lr){12-14} & Angle$\degree$ $\downarrow$ & PSNR $\uparrow$ & SSIM $\uparrow$ & LPIPS $\downarrow$ & PSNR $\uparrow$ & SSIM $\uparrow$ & LPIPS $\downarrow$ & PSNR $\uparrow$ & SSIM $\uparrow$ & LPIPS $\downarrow$ & PSNR $\uparrow$ & SSIM $\uparrow$ & LPIPS $\downarrow$ \\
        \midrule
        SIRFS & - & 26.0204 & 0.9420 & 0.0719 & - & - & - & - & - & - & - & - & - \\
        Oxholm \& Nishino\textdagger & 32.0104 & 26.3248 & 0.9448 & 0.0870 & 29.8093 & 0.9275 & 0.0810 & 20.9979 & 0.8407 & 0.1610 & 22.2783 & 0.8762 & 0.1364 \\
        \midrule

\model & \cellcolor{JonRed}{22.1327} & \cellcolor{JonOrange}{28.7099} & \cellcolor{JonOrange}{0.9533} & \cellcolor{JonOrange}{0.0621} & \cellcolor{JonOrange}{32.5362} & \cellcolor{JonRed}{0.9461} & \cellcolor{JonOrange}{0.0457} & \cellcolor{JonOrange}{23.6206} & \cellcolor{JonRed}{0.8647} & \cellcolor{JonOrange}{0.1264} & \cellcolor{JonRed}{26.6275} & \cellcolor{JonRed}{0.9026} & \cellcolor{JonRed}{0.0917}  \\
\modelmicrofacet & \cellcolor{JonOrange}{22.1804} & \cellcolor{JonRed}{29.1608} & \cellcolor{JonRed}{0.9571} & \cellcolor{JonRed}{0.0567} & \cellcolor{JonYellow}{32.4409} & \cellcolor{JonOrange}{0.9457} & \cellcolor{JonYellow}{0.0458} & \cellcolor{JonRed}{23.7885} & \cellcolor{JonOrange}{0.8642} & \cellcolor{JonRed}{0.1256} & \cellcolor{JonOrange}{26.5970} & \cellcolor{JonOrange}{0.9011} & \cellcolor{JonYellow}{0.0925}  \\
w/o geom. pretrain. & \cellcolor{JonYellow}{25.5302} & 27.7936 & \cellcolor{JonYellow}{0.9480} & \cellcolor{JonYellow}{0.0677} & 32.3835 & 0.9449 & 0.0491 & \cellcolor{JonYellow}{23.1689} & \cellcolor{JonYellow}{0.8585} & 0.1384 & 25.8185 & \cellcolor{JonYellow}{0.8966} & 0.1027  \\
w/o smoothness & 26.2229 & 27.7389 & 0.9179 & 0.0853 & \cellcolor{JonRed}{32.7156} & \cellcolor{JonYellow}{0.9450} & \cellcolor{JonRed}{0.0405} & 23.0119 & 0.8455 & \cellcolor{JonYellow}{0.1283} & \cellcolor{JonYellow}{26.0416} & 0.8887 & \cellcolor{JonOrange}{0.0920}  \\
\modelnerfshape & 32.0634 & \cellcolor{JonYellow}{27.8183} & 0.9419 & 0.0689 & 30.7022 & 0.9210 & 0.0614 & 22.0181 & 0.8237 & 0.1470 & 24.8908 & 0.8651 & 0.1154  \\

        \bottomrule
    \end{tabular}
    }
    \begin{flushleft}\footnotesize
    \textdagger\citet{oxholm2014multiview} requires the ground-truth illumination, which we provide, and this baseline represents a significantly enhanced version (see \sect{sec:baselines}).
    \end{flushleft}
\end{table*} 
\begin{figure*}[!htbp]
  \centering
  \includegraphics[width=\textwidth]{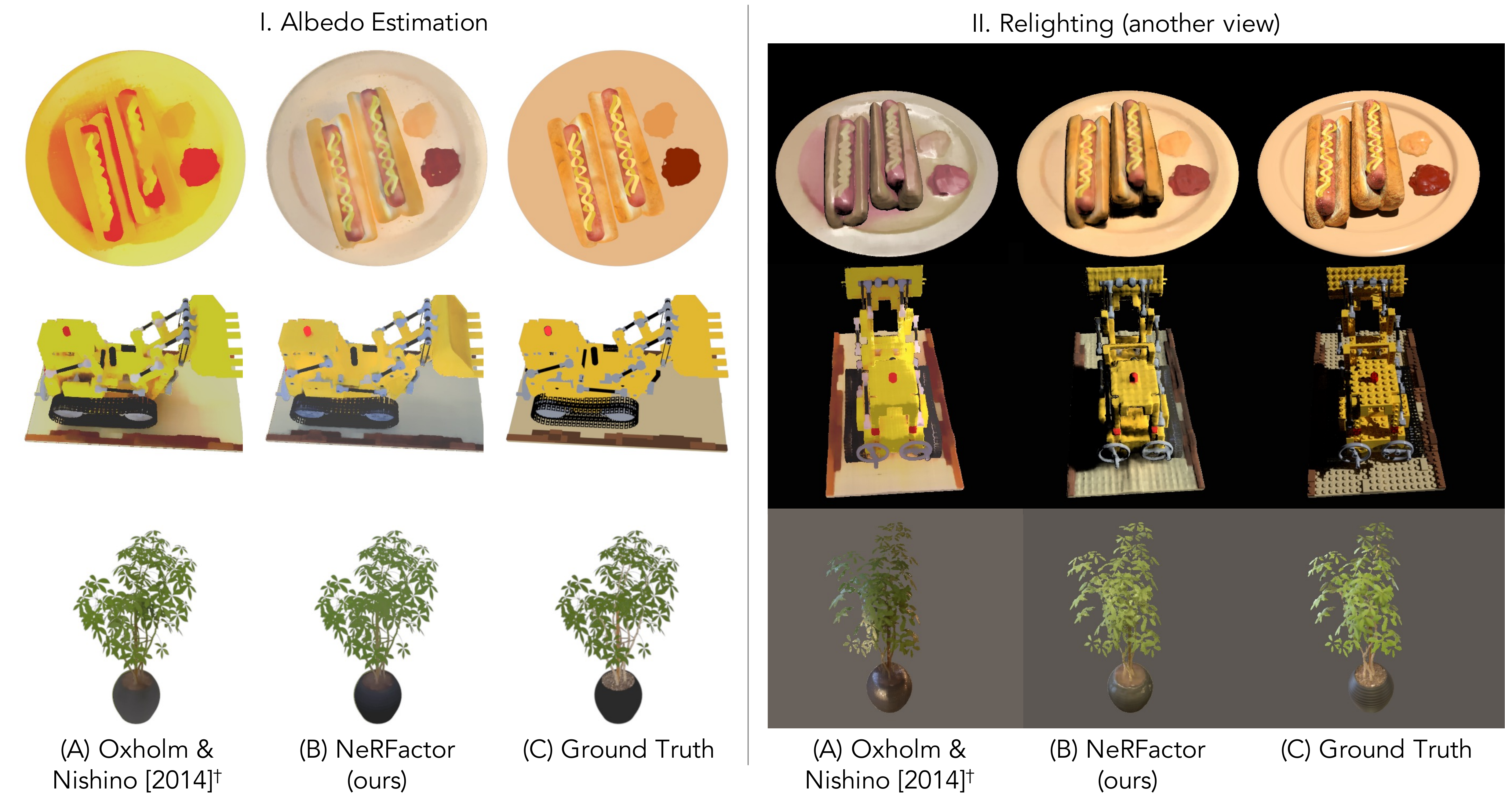}
  \vspace{-6ex}
  \caption{
  \textbf{Comparisons against \citet{oxholm2014multiview}.}
  See \sect{sec:baselines} for discussions.
  \textdagger We significantly enhanced this baseline, as explained in \sect{sec:baselines}; in addition, we provide it with the ground-truth illumination since it does not estimate lighting.
  }
  \label{fig:oxholm}
\end{figure*} 
\paragraph{Learned \vs analytic BRDFs}
\rev{
Instead of using an MLP to parametrize the BRDF and pretraining it on an external BRDF dataset to learn data-driven priors, one can adopt an analytic BRDF model such as the microfacet model of \citet{walter2007microfacet} and ask an MLP to predict spatially-varying roughness for the microfacet BRDF.
As \tbl{tbl:quant} shows, this model variant achieves good performance across all tasks but overall underperforms \model.
Note that to improve this variant, we remove the smoothness constraint on the predicted roughness because even a tiny smoothness weight still drives the optimization to the local optimum of predicting maximum roughness everywhere (this local optimum is a ``safe'' solution that renders everything more diffuse to satisfy the $\ell^2$ reconstruction loss).
As such, this model variance sometimes produces noisy rendering due to its non-smooth BRDFs as shown in \suppvideo.
}

\paragraph{With \vs without geometry pretraining}
As shown in \fig{fig:model_a} and discussed in \sect{sec:method}, we pretrain the normal and visibility MLPs to just reproduce the NeRF values given $\bm{x_\text{surf}}$ before plugging them into the joint optimization (where they are then fine-tuned together with the rest of the pipeline), to prevent the albedo MLP from mistakenly attempting to explain way the shadows.
Alternatively, one can train these two geometry MLPs from scratch together with the pipeline.
As \tbl{tbl:quant} shows, this variant indeed predicts worse albedo with shading residuals (\fig{fig:ablation} [C]) and overall underperforms \model.

\paragraph{With \vs without smoothness constraints} 
In \sect{sec:method}, we introduce our simple yet effective spatial smoothness constraints in the context of MLPs and their crucial role in this underconstrained setup.
Ablating these smoothness constraints does not prevent this variant from performing well on view synthesis (similar to how NeRF is capable of high-quality view synthesis without any smoothness constraints) as shown in \tbl{tbl:quant}, but does hurt this variant's performance on other tasks such as albedo estimation and relighting.
Qualitatively, this variant produces noisy estimations insufficient for relighting (see \fig{fig:ablation} [B] and \suppvideo).

\paragraph{Optimizing shape \vs just using NeRF's shape}
If we ablate the normal and visibility MLPs entirely, this variant is essentially using NeRF's normals and visibility without improving upon them (hence ``\modelnerfshape'').
As \tbl{tbl:quant} and \suppvideo show, even though the estimated reflectance is smooth (encouraged by the smoothness priors from the full model), the noisy NeRF normals and visibility produce artifacts in the final rendering.

\subsection{Baseline Comparisons}
\label{sec:baselines}

In this section, we compare \model with both classic and deep learning-based state of the art (\citet{oxholm2014multiview} and \citet{philip2019multi}) in the tasks of appearance factorization and free-viewpoint relighting.

\rev{
See \sect{sec:morebaselines} of the appendix for comparisons with the classic single-view SIRFS approach \cite{Barron2015Shape}.
}

\paragraph{\citet{oxholm2014multiview}}
We compare \model with a significantly improved version of the multi-view approach that estimates the shape and \emph{non-}spatially-varying BRDF under a \emph{known} lighting condition \cite{oxholm2014multiview}.
Due to the source code being unavailable, we re-implemented this method in our framework, capturing the main ideas of smoothness regularization on shape and data-driven BRDF priors, and then enhanced it with a better shape initialization (visual hull \textrightarrow \ NeRF shape) and the ability to model spatially-varying albedo (the original paper considers only non-spatially-varying BRDFs).
Other differences include representing the shape with a surface normal MLP instead of mesh and expressing the predicted BRDF with a pretrained BRDF MLP instead of MERL BRDF bases \cite{nishino2009directional,nishino2011directional,lombardi2012reflectance}. Also note that this baseline has the advantage of receiving the ground-truth lighting as input, whereas \model has to estimate lighting together with shape and reflectance.

As shown in \fig{fig:oxholm} (I), even though this improved version of \citet{oxholm2014multiview} has access to the ground-truth illumination, it struggles to remove shadow residuals from the albedo estimation because of its inability to model visibility (\texttt{hotdog} and \texttt{lego}).
As expected, these residuals in albedo negatively affect the relighting results in \fig{fig:oxholm} (II) (\eg, the red shade on the \texttt{hotdog} plate).
Moreover, because the BRDF estimated by this baseline is not spatially-varying, BRDFs of the hot dog buns and the ficus leaves are incorrectly estimated to be as specular as the plate and vase, respectively.
Finally, this baseline is unable to synthesize non-local light transport effects such as shadows (\texttt{hotdog} and \texttt{lego}), in contrast to \model that correctly produces realistic hard cast shadows under the OLAT conditions.

\paragraph{\citet{philip2019multi}}
The recent work by \citet{philip2019multi} presents a technique to relight large-scale scenes and specifically focuses on synthesizing realistic shadows.
The input to their system is similar to ours: multi-view images of a scene lit by unknown lighting.
However, their technique only supports synthesizing images illuminated by a single primary light source such as the Sun.
In other words, unlike \model, their approach does not support relighting with arbitrary lighting such as another random light probe.
As such, we compare it with \model only on the task of point light relighting.

\begin{figure}[!htbp]
  \centering
  \includegraphics[width=\columnwidth]{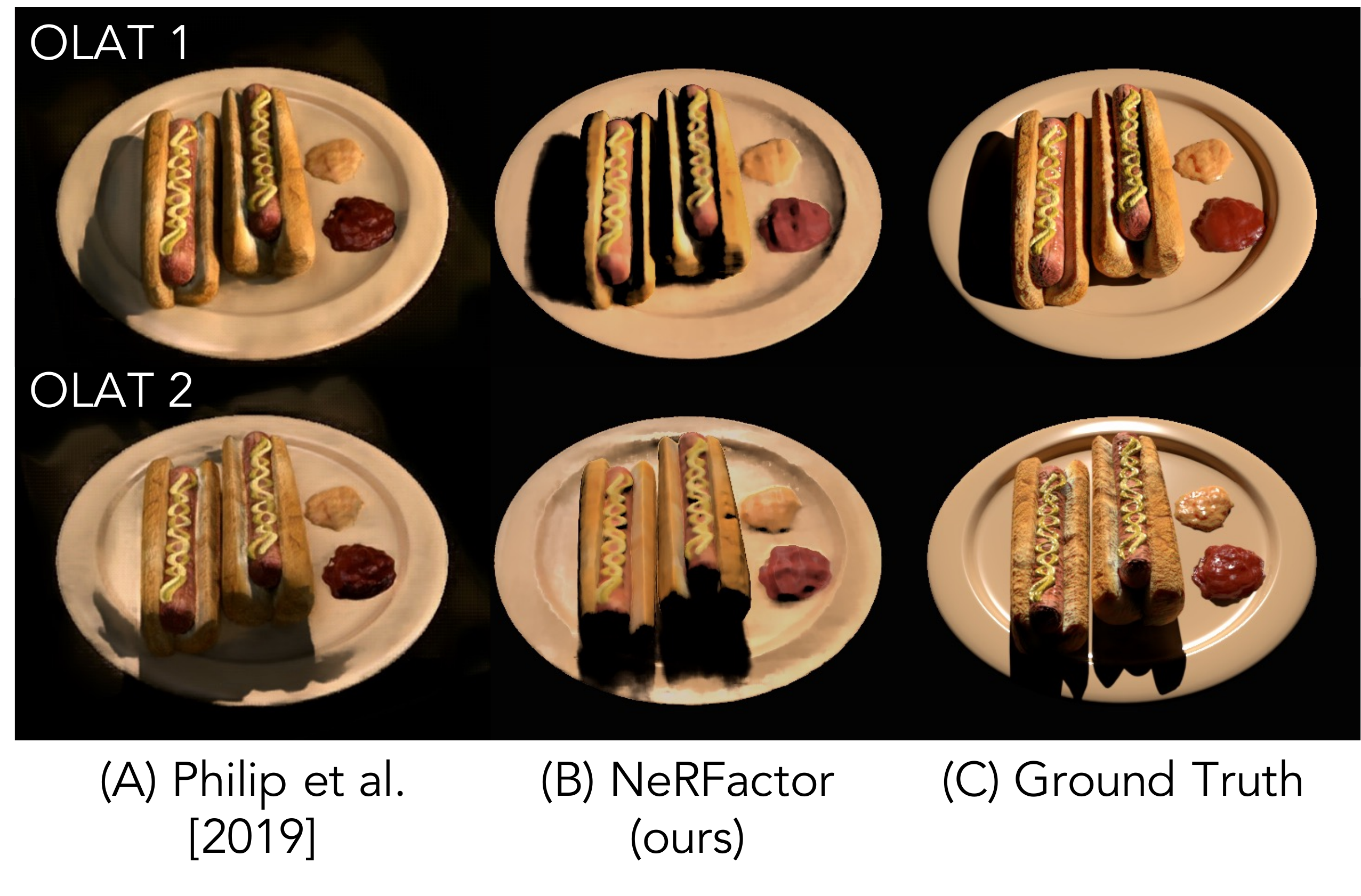}
  \begin{tabular}{lccc}
        \toprule
         & PSNR $\uparrow$ & SSIM $\uparrow$ & LPIPS $\downarrow$ \\
        \midrule
        \citet{philip2019multi} & 20.0397 & 0.5000 & 0.1812 \\
        \citet{philip2019multi} + Masks & 21.8620 & 0.8436 & \textbf{0.1140} \\
        NeRFactor (ours) & \textbf{22.9625} & \textbf{0.8592} & 0.1230 \\
        \bottomrule
  \end{tabular}
  \vspace{-1ex}
  \caption{
  \textbf{Comparisons with \citet{philip2019multi} in point light relighting.}
  The numbers here are averages over eight test OLAT conditions.
  See \sect{sec:baselines} for discussions.
  }
  \label{fig:philip}
\end{figure}

The ``yellow fog'' in the background of their results (\fig{fig:philip} [A]) is likely due to the poor geometry reconstruction.
Because their network is trained on outdoor scenes (not images with backgrounds), we additionally compute error metrics after masking out the yellow fog with the ground-truth object masks (``\citet{philip2019multi} + Masks'') for a more generous comparison.
As the table in \fig{fig:philip} shows, \model outperforms ``\citet{philip2019multi} + Masks'' in both PSNR and SSIM.
The baseline achieves a lower (better) LPIPS score because it renders new viewpoints by reprojecting observed images using estimated proxy geometry, as is typical of Image-Based Rendering (IBR) algorithms.
Thus, it retains the high-frequency details present in the input images, resulting in a lower LPIPS score.
However, as a physically-based (re-)rendering approach that operates fully in the 3D space, \model synthesizes shadows that better match the ground truth (while the baseline's shadows tend to be overly soft [OLAT 1] or cover a less accurate region [OLAT 2]) and supports relighting with arbitrary light probes such as ``Studio,'' which has four major light sources (\fig{fig:real} [D]).
\section{Limitations}

Although we demonstrate that \model outperforms its variants and the baseline methods, there are still a few important limitations.
First, to keep light visibility computation tractable, we limit the resolution of the light probe images to $16\times32$, a resolution that may be insufficient for generating very hard shadows or recovering very high-frequency BRDFs.
As such, when the object is lit by a very high-frequency illumination such as the one in \fig{fig:consistency} (Case D) where the sun pixels are fully HDR, there might be specularity or shadow residuals in the albedo estimation (such as those on the vase).
Second, for fast rendering, we consider only single-bounce direct illumination, so \model does not properly account for indirect illumination effects. %
Finally, \model initializes its geometry estimation with NeRF or MVS.
While it is able to fix errors made by NeRF up to a certain degree, it can fail if NeRF estimates particularly poor geometry in a manner that happens to not affect view synthesis.
We observe this in the two real-world NeRF scenes, which contain faraway incorrect ``floating'' geometry that is not visible from the input cameras but casts shadows on the objects.  %
\section{Conclusion}

In this paper, we have presented \modelfull, a method that recovers an object's shape and reflectance from multi-view images and their camera poses.
Importantly, \model recovers these properties from images under an unknown illumination condition, while the majority of prior work requires observations under multiple known illumination conditions.
To address the ill-posed nature of this problem, \model relies on priors to estimate a set of plausible shape, reflectance, and lighting that collectively explain the observed images.
These priors include simple yet effective spatial smoothness constraints (implemented in the context of Multi-Layer Perceptrons [MLPs]) and a data-driven prior on real-world BRDFs.
We demonstrate that \model achieves high-quality geometry sufficient for relighting and view synthesis, produces convincing albedo as well as spatially-varying BRDFs, and generates lighting estimations that correctly reflect the presence or absence of dominant light sources.
With \model's factorization, we can relight the object with point lights or light probe images, render images from arbitrary viewpoints, and even edit the object's albedo and BRDF.
We believe that this work makes important progress towards the goal of recovering fully-featured 3D graphics assets from casually-captured photos.

\begin{acks}
We thank
the anonymous reviewers for their helpful suggestions,
Julien Philip, Tiancheng Sun, Zhang Chen, Jonathan Dupuy, and Wenzel Jakob for their help in providing their data or results,
Zhoutong Zhang, Xuaner (Cecilia) Zhang, Yun-Tai Tsai, Jiawen Chen, Tzu-Mao Li, Yonglong Tian, and Noah Snavely for fruitful discussions,
Noa Glaser and David Salesin for their constructive comments on the paper.
This work was partially funded by the MIT-Air Force AI Accelerator.
We thank the following \url{blendswap.com} users for the models used as our synthetic scenes: \texttt{bryanajones} (\texttt{drums}), \texttt{Herberhold} (\texttt{ficus}), \texttt{erickfree} (\texttt{hotdog}), and \texttt{Heinzelnisse} (\texttt{lego}).
\end{acks} 
\bibliographystyle{ACM-Reference-Format}
\bibliography{main}

\appendix

\renewcommand{\thefigure}{S\arabic{figure}}
\setcounter{figure}{0}

\rev{
\vspace{4ex}
\begin{flushleft}\LARGE
\textbf{Supplemental Information}
\end{flushleft}
}

\section{Implementation Details}
\label{sec:impl}

\model is implemented in TensorFlow 2 \cite{abadi2016tensorflow}.
All training uses the Adam optimizer \cite{Kingma2015Adam} with the default hyperparameters.
See \repo for our implementation that reproduces the results in this paper.

\subsection{Staged Training}

There are three stages in training \model.
First, we optimize a NeRF using the input images and their camera poses (once per scene), and train a BRDF MLP on the MERL dataset (only once for all scenes).
Both of these MLPs are frozen during the final joint optimization since the NeRF only provides a shape initialization, and the BRDF MLP provides a latent space of real-world BRDFs for the optimization to explore.
Future shape refinement happens in \model's normal and visibility MLPs, and the actual material prediction happens in \model's albedo and BRDF identity MLPs.
Second, we use this trained NeRF to initialize our geometry by optimizing the normal and visibility MLPs to simply reproduce the NeRF values, without any additional smoothness loss or regularization.
Finally, we jointly optimize the albedo MLP, BRDF identity MLP, and light probe pixels from scratch, along with the pretrained normal and visibility MLPs.
Finetuning the normal and visibility MLPs along with the reflectance and lighting allows the errors in NeRF's initial geometry to be fixed in order to minimize the re-rendering loss (\fig{fig:shape-results}).

\subsection{Architecture \& Positional Encoding}

We use the default architecture for NeRF \cite{mildenhall2020nerf}, and all other MLPs we introduce contain four layers (with a skip connection from the input to the second layer), each with $128$ hidden units.
As in NeRF \citep{mildenhall2020nerf}, we apply positional encoding to the input coordinates of all networks with $10$ encoding levels for 3D locations and $4$ encoding levels for directions.

\subsection{Runtime}

We train NeRF for \num{2000} epochs, which takes $6$--\SI{8}{\hour} when distributed over four NVIDIA TITAN RTX GPUs.
Prior to the final joint optimization, computing the initial surface normals and light visibility from the trained NeRF takes \SI{30}{\minute} per view on one GPU for a $16\times32$ light probe (\ie, $512$ light locations).
This step can be trivially parallelized because each view is processed independently.
Geometry pretraining is performed for $200$ epochs, which takes around \SI{20}{\minute} on a TITAN RTX.
The final joint optimization is performed for $100$ epochs, which takes only \SI{30}{\minute} on one TITAN RTX.

\section{Data}
\label{sec:data}

This work uses three types of data: multi-view images of an object and the corresponding camera poses, real-world measured BRDFs, and captured light probes.

\subsection{Synthetic Renderings}

We use the synthetic Blender scenes (\texttt{hotdog}, \texttt{drums}, \texttt{lego}, and \texttt{ficus}) released by \citet{mildenhall2020nerf} and replace the lighting used there with our own natural illuminations taken from real light probe images.
The light probes are from \url{hdrihaven.com}, \citet{stumpfel2006direct}, and the Blender distribution%
\footnote{\url{https://www.blender.org}}.
This yields significantly more natural input illumination conditions.
We also disable all non-standard post-rendering effects used by Blender Cycles when rendering the images, such as ``filmic'' tone mapping, and retain only the standard linear-to-sRGB tone mapping.
We render all images directly to PNGs instead of EXRs to simulate real-world mobile phone captures where raw HDR pixel intensities may not be available; this indeed facilitates applying \model directly to real scenes as shown in \fig{fig:real} and \fig{fig:mvs}.

\subsection{Real Captures}

We use mobile phone captures of two real scenes released by \citet{mildenhall2020nerf}: \texttt{vasedeck} and \texttt{pinecone}.
These scenes are captured by inwards-facing cameras on the upper hemisphere.
There are close to $100$ images per scene, and the camera poses are obtained by COLMAP SFM \cite{schoenberger2016sfm}.
\model is directly applicable because it is designed to work with PNGs instead of EXRs.

\rev{
In addition, we also use real images from the DTU-MVS dataset \cite{jensen2014large,aanaes2016large}.
Each scene in this dataset consists of around $50$ multi-view images and their corresponding camera poses for each scene.
We use images under ``the most diffuse lighting'' in DTU-MVS for all scenes%
\footnote{These are the ``\texttt{*\_3\_*}'' images in the DTU-MVS release.}.
The 3D surfaces are from the Poisson reconstruction \cite{Kazhdan2006Poisson} of the MVS reconstruction by \citet{furukawa2009accurate}, as bundled in DTU-MVS%
\footnote{These are the ``\texttt{furu???\_l3\_surf\_11\_trim\_8.ply}'' mesh files in their release.}.
}

\subsection{Measured BRDFs}

We use real measured BRDFs from the MERL dataset by \citet{matusik2003}.
The MERL dataset consists of $100$ real-world BRDFs measured by a conventional gonioreflectometer.
Because the color components of BRDFs are not used by our model, we convert the RGB reflectance values to be achromatic by converting linear RGB values to relative luminance.

\section{Additional Evaluation Studies}

Here we study whether the albedo estimation by \model is consistent for the same object when lit by different input lighting conditions.
We then visualize the importance of different model components to supplement the quantitative ablation studies in \tbl{tbl:quant}.
Finally, we compare \model with SIRFS, a classic single-view approach by \citet{Barron2015Shape}.

\subsection{Estimation Consistency Across Different Illuminations}
\label{sec:consistency}

\begin{figure}[!htbp]
  \centering
  \includegraphics[width=\columnwidth]{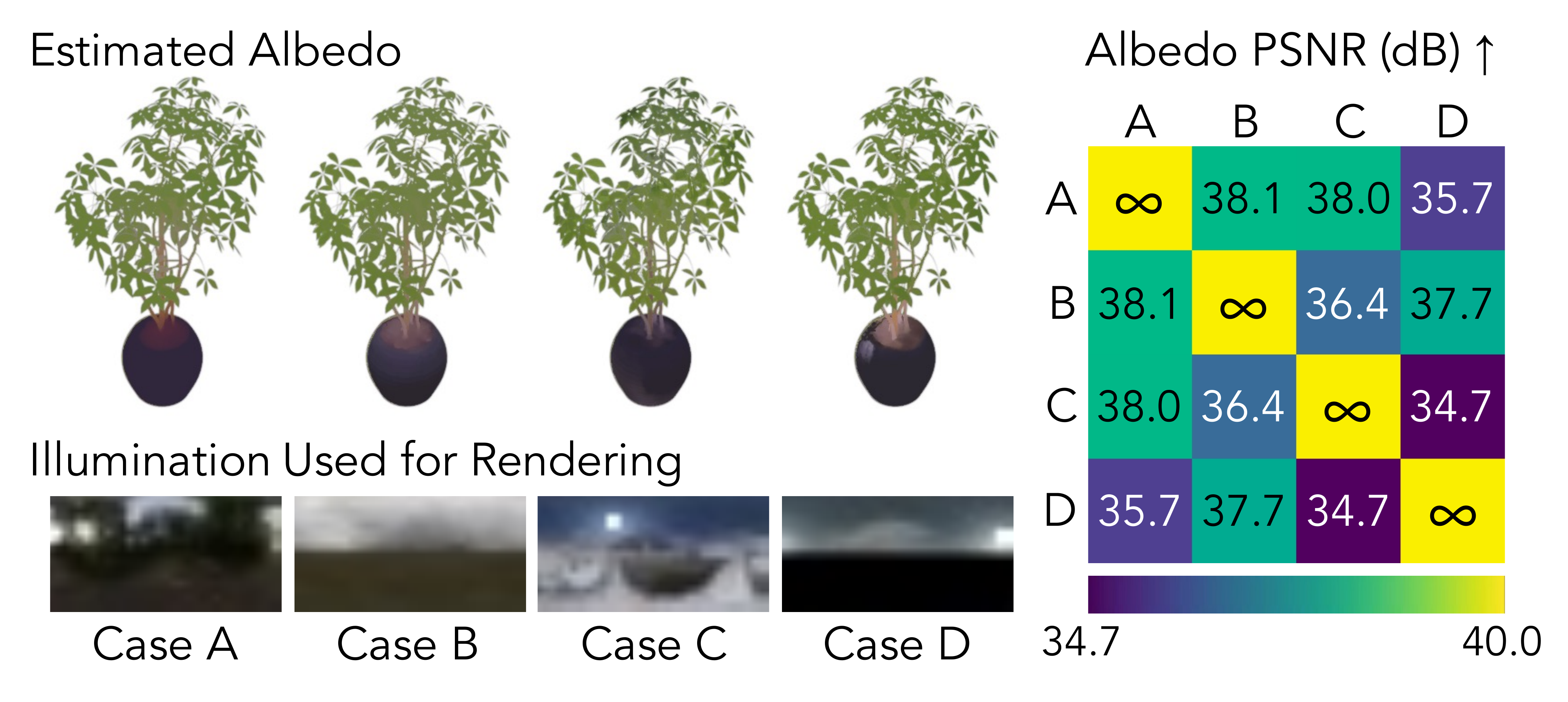}
  \vspace{-6ex}
  \caption{
  \textbf{Albedo estimation consistency across different input illumination conditions.}
  The albedo fields recovered by \model are largely consistent across varying illumination conditions of the input images.
  }
  \label{fig:consistency}
\end{figure}

In this experiment, we study how different illumination conditions affect the albedo estimation by \model.
More specifically, we probe how consistent the estimated albedo predictions are across different input illumination conditions.
To this end, we light \texttt{ficus} with four drastically different lighting conditions, as shown in \fig{fig:consistency}, and then estimate the albedo with \model from these four sets of multi-view images.

As \fig{fig:consistency} shows, \model's predictions are similar across the four input illuminations, with pairwise PSNR $\geq 34.7\ \text{dB}$.
Note that the performance on Case D is worse (\eg, the specularity residuals on the vase) than on Case C, despite that both cases seem to have the Sun as the primary light source.
The reason is that Case D has the Sun pixels properly measured by \citet{stumpfel2006direct}, whereas Case C is an internet light probe that clips the Sun pixels.
Therefore, Case D has a much higher-frequency lighting condition than Case C, making it a harder case for \model to correctly factorize the appearance.

\subsection{Qualitative Ablation Studies}
\label{sec:qualiablation}

\begin{figure*}[!htbp]
  \centering
  \includegraphics[width=\textwidth]{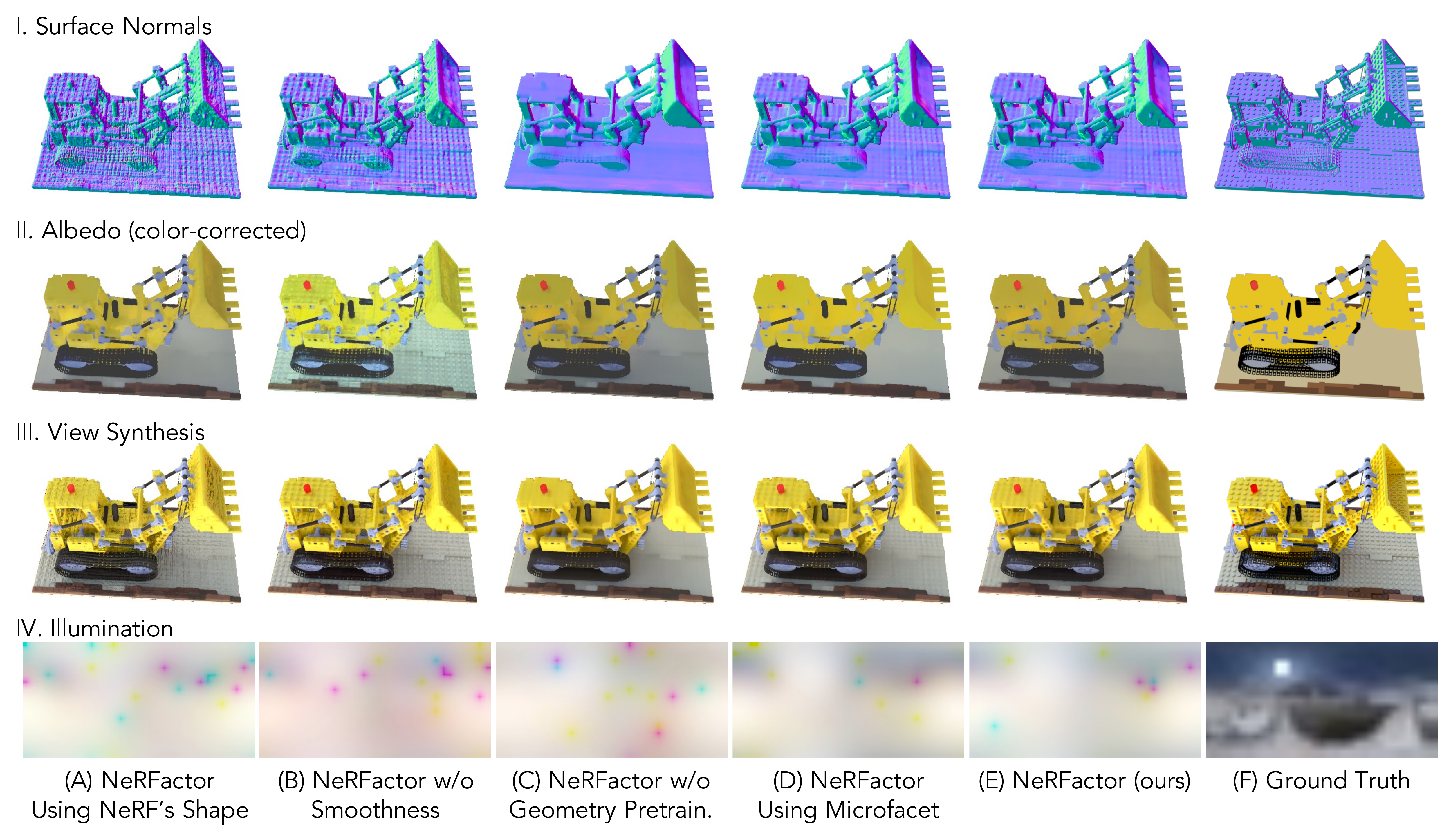}
  \vspace{-2em}
  \caption{
  \textbf{Ablation studies.}
  See \sect{sec:qualiablation} for discussions.
  }
  \label{fig:ablation}
\end{figure*}

\fig{fig:ablation} shows, qualitatively, what happens when each of the major model components is ablated.
See \sect{sec:ablation} for more discussions and \tbl{tbl:quant} for the quantitative evaluation.

As \fig{fig:ablation} (A) shows, one can fix the geometry to that of NeRF and estimate only the reflectance and illumination by ablating the normal and visibility MLPs of \model, but the NeRF geometry is too noisy (I) to be used for relighting (see \suppvideo).
(B) Ablating the smoothness regularization leads to noisy geometry and albedo (I, II).
(C) If we train the normal and visibility MLPs from scratch during the joint optimization (\ie, no pretraining), the recovered albedo may mistakenly attempt to explain shading and shadows (III).  
(D) If we replace the learned BRDF with an MLP predicting the roughness parameter of a microfacet BRDF, the predicted reflectance either falls into the local optimum of maximum roughness everywhere or becomes spatially non-smooth (not pictured here; see \suppvideo).
(E) \model is able to recover a \emph{plausible} set of normals, albedo, and lighting without direct supervision on any factor.
The lighting recovered by \model, though oversmoothed, correctly captures the location of the Sun.

\subsection{More Baseline Comparisons}
\label{sec:morebaselines}

In addition to \citet{oxholm2014multiview} and \citet{philip2019multi} (\sect{sec:baselines}), here we also compare \model with SIRFS \cite{Barron2015Shape}, both qualitatively and quantitatively.

\begin{figure*}[hbpt]
  \centering
  \includegraphics[width=\textwidth]{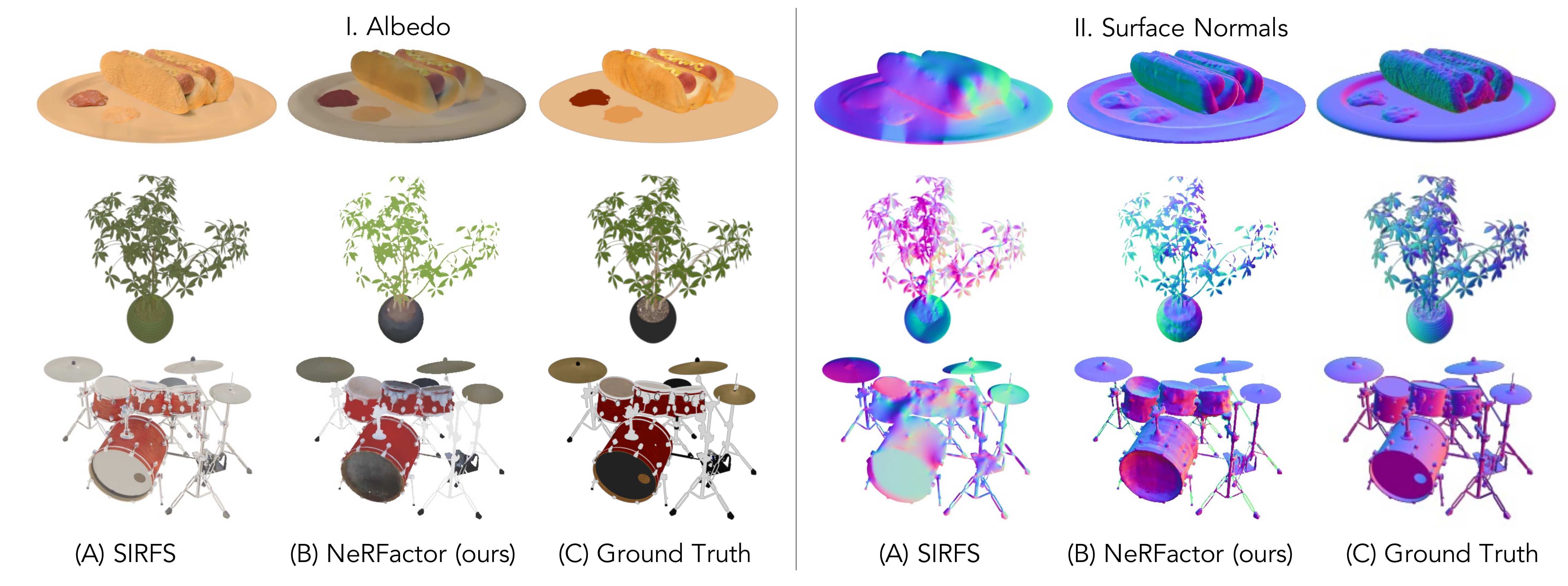}
  \vspace{-6ex}
  \caption{
  \textbf{Comparisons with SIRFS.}
  Although the albedo estimation by SIRFS is reasonable, the surface normals are highly inaccurate (likely due to its inability to use multiple images to inform shape estimation).
  }
  \label{fig:sirfs}
\end{figure*}

SIRFS is a single-image method that decomposes appearance into surface normals, albedo, and shading (not shadowing) in the input view under unknown lighting.
In contrast, \model is a multi-view approach that estimates these properties plus BRDFs and visibility (hence, shadows) in the full 3D space alongside the unknown lighting.
In other words, \model gets to observe many more views than SIRFS, which observes only one view.
Under this setup, \model outperforms SIRFS quantitatively as shown by \tbl{tbl:quant}.
\fig{fig:sirfs} shows that although SIRFS achieves reasonable albedo estimation, it produces inaccurate surface normals likely due to its inability to incorporate multiple views or to reason about shape in ``world space.''
In addition, SIRFS is unable to render the scene from arbitrary viewpoints or synthesize shadows during relighting.

\end{document}